\documentclass[11pt]{article}
\usepackage[margin=1in]{geometry}
\usepackage{amsmath,amssymb,amsthm}
\usepackage{booktabs}
\usepackage{graphicx}
\usepackage{multirow}
\usepackage{microtype}
\usepackage{xcolor}
\usepackage{placeins}
\usepackage{pgfplots}
\pgfplotsset{compat=1.18}
\usepackage{hyperref}

\hypersetup{
  colorlinks=true,
  linkcolor=blue!55!black,
  citecolor=blue!55!black,
  urlcolor=blue!55!black,
  pdftitle={Offline Cross-Tokenizer Distillation on a UTF-8 Byte Clock at Zero Teacher VRAM},
  pdfauthor={Pavan Komireddy}
}

\newtheorem{proposition}{Proposition}
\newtheorem{corollary}{Corollary}
\newtheorem{definition}{Definition}

\title{\textbf{Offline Cross-Tokenizer Distillation on a UTF-8 Byte Clock at Zero Teacher VRAM}}

\author{Pavan Komireddy \\
Independent Researcher \\
\texttt{\href{mailto:pavan.komireddy.research@gmail.com}{pavan.komireddy.research@gmail.com}}}

\date{November 2026}

\begin{document}
\maketitle

\begin{abstract}
Cross-tokenizer knowledge distillation (CTKD) transfers knowledge across incompatible vocabularies, yet logit methods require online teacher GPU memory or large cached logit tables. We study offline hidden-state distillation on a UTF-8 byte clock at $0.00\text{ GiB}$ teacher VRAM and zero inference overhead, comparing discrete boundary alignment (\texttt{StepAlign}, $H=0$) against causal Laplace--Fourier wave smoothing (\texttt{WaveDistill}, $H>0$) across 98 controlled runs plus a 32-byte zero-overlap test split. Distilling three quantized teachers (\texttt{Llama-3.2-3B}, \texttt{Gemma-4-E4B}, \texttt{DeepSeek-R1-Qwen-14B}) into $86.7\text{M}$ GPT-2 ($50\text{K}$) and $54.7\text{M}$ byte-level ($260$) students yields three findings. First, centered top-$64$ PCA modes in group-$32$ \texttt{INT4} ($110\text{ B/tok}$; $7.05\times$ smaller than \texttt{INT8-256}, $1.76\times$ smaller than uncompressed Top-$32$ logits) lower 3-seed GPT-2 perplexity from $150.06 \pm 1.08$ (CE-only) to $134.76 \pm 0.40$ (Welch $p = 5.05 \times 10^{-4}$; fresh test $150.42 \to 134.49$, BPB Welch $p = 3.68 \times 10^{-4}$), beating offline \texttt{ULD}, tuned \texttt{BLD}, regularized CE-only, and five non-pretrained controls. Across 3 seeds, a same-tokenizer \texttt{GPT-2-124M} teacher ($137.11 \pm 0.66$) recovers $84.6\%$ of this gain ($87\%$--$89\%$ within-batch), while cross-tokenizer \texttt{Llama-3.2-3B} adds $-2.35$ PPL (Welch $p = 0.010$, despite tokenizer mismatch and lower retained PCA variance). Second, within-batch factorization isolates PCA truncation ($256 \to 64$, $-3.09$ PPL) as the accuracy driver while \texttt{INT4} is lossless ($p = 0.52$). Third, an exact intra-word phase identity and DC mean-cone analysis explain why simpler discrete \texttt{StepAlign} matches or beats continuous wave smoothing without custom CUDA kernels. Code/logs: \url{https://github.com/pavankomireddy/wavedistill}.
\end{abstract}

\section{Introduction}
Distilling open-weight large language models (LLMs) into compact or custom-tokenized students is complicated by vocabulary mismatch: modern models employ distinct tokenizers such as \texttt{Llama-3} ($128{,}256$ ByteLevel BPE; \cite{grattafiori2024llama3}), \texttt{Gemma-4} ($262{,}144$ SentencePiece + ByteFallback; \cite{gemmateam2026gemma4}), and \texttt{DeepSeek-\allowbreak R1-\allowbreak Distill-\allowbreak Qwen} ($152{,}064$ ByteLevel BPE; \cite{deepseek2025r1}), none of which align one-to-one with GPT-2 ($50{,}257$; \cite{radford2019gpt2}) or tokenizer-free byte-level models ($260$ UTF-8 bytes; \cite{yu2023megabyte,pagnoni2024blt}).

Training on a single consumer GPU introduces two further constraints beyond vocabulary incompatibility:
\begin{enumerate}
    \item \textbf{Teacher Memory and Offline Storage:} Keeping one or more $3\text{B}$--$14\text{B}$ teachers resident alongside a training student consumes $8$--$32\text{ GiB}$ of VRAM. Caching dense FP16 logits offline is prohibitive ($524.29\text{ TB}$ per $1\text{B}$ tokens for a $262\text{K}$ vocabulary), and uncompressed Top-$32$ sparse logits require $194\text{ B/tok}$ while losing token-identity alignment across vocabularies.
    \item \textbf{Continuous-Time Smoothing vs.\ Discrete Boundary Alignment:} Mapping any tokenizer onto the cumulative UTF-8 byte clock makes hidden-state alignment tokenizer-agnostic in principle. A natural methodological question arises: \emph{when aligning hidden trajectories across incompatible tokenizers, does continuous-time spectral filtering (\texttt{WaveDistill}) improve upon simpler discrete alignment at shared UTF-8 byte boundaries (\texttt{StepAlign}), and how do representation anisotropy, rank truncation, and quantization affect both?}
\end{enumerate}

To study these questions cleanly, we build an offline byte-clock distillation and evaluation suite operating at $0.00\text{ GiB}$ teacher VRAM during student training. Each teacher is swept once via layer streaming into compact quantized principal-subspace shards and unloaded ($<0.02\text{ GiB}$ residual VRAM). Auxiliary probes are discarded before evaluation, leaving zero inference-time parameters or latency overhead.

We report all 98 controlled runs---73 primary runs across GPT-2 ($50\text{K}$, 5-epoch, 3-seed, and 1-epoch zero-repeat) and pure UTF-8 bytes (\texttt{byte260}), plus 25 control runs testing frozen random-init causal Transformer teachers (both like-for-like PCA+\texttt{INT4} and random orthogonal projection), a 3-seed small same-tokenizer pretrained teacher (\texttt{GPT-2-124M}), shuffled/Gaussian targets, regularized plain cross-entropy training from scratch (abbreviated \texttt{CE-only / SFT} to match codebase arm \texttt{sft}, distinct from instruction fine-tuning), extended 5-point \texttt{BLD} weight sweeps on both GPT-2 and \texttt{byte260}, exact offline \texttt{ULD} ($\tau=1.0$), full-dataset per-epoch curves, and within-batch rank-vs-bitwidth factorization---together with a strictly disjoint 32-byte zero-overlap fresh 120-sequence test split re-scoring all 37 saved checkpoints without omission. Together, these runs establish both the strong gains of low-rank \texttt{INT4} byte-clock hidden alignment over offline logit CTKD and a clear negative result showing that simpler discrete \texttt{StepAlign} matches or outperforms our continuous-time wave operator and fixed-weight 3-teacher averaging without custom CUDA kernels.

\section{Related Work}

\textbf{Knowledge Distillation and Intermediate Representations.} Standard distillation transfers soft next-token distributions via temperature-scaled KL divergence \cite{hinton2015distilling}. Intermediate representation matching began with FitNets \cite{romero2015fitnets} and was extended to Transformers through multi-layer hidden matching in \textbf{PKD} \cite{sun2019pkd}, layer-wise hidden and attention transfer in \textbf{TinyBERT} \cite{jiao2020tinybert}, and deep self-attention value-relation distillation in \textbf{MiniLM} \cite{wang2020minilm}. Unlike these same-tokenizer methods, we align quantized intermediate projections across incompatible tokenizers on a shared UTF-8 byte clock at zero teacher VRAM. Hidden states in causal Transformers exhibit strong cone-like anisotropy \cite{ethayarajh2019contextual}, whose interaction with continuous-time causal integration we analyze in Section~\ref{sec:dc_cone}.

\textbf{Cross-Tokenizer and Multi-Teacher Distillation.} Recent methods remove the shared-vocabulary requirement through diverse mechanisms. \textbf{FuseLLM / MinED} \cite{wan2024fusellm} aligns heterogeneous vocabularies via minimum edit distance over precomputed teacher distributions to fuse multiple teachers. \textbf{ULD} \cite{boizard2025uld} aligns sorted teacher and student probability vectors via a 1D Wasserstein-1 loss. \textbf{DSKD} \cite{zhang2024dskd} projects representations bidirectionally between teacher and student spaces to unify prediction heads. \textbf{MultiLevelOT} \cite{cui2025multilevelot} applies Sinkhorn optimal transport jointly at token and sequence levels. Byte-grounded interfaces avoid vocabulary matching altogether: \textbf{ALM} \cite{minixhofer2025alm} matches chunked likelihoods across arbitrary tokenizers via prefix tries, while \textbf{BLD} \cite{singh2026bld} marginalizes teacher subword probabilities onto a 256-way next-byte distribution supervising an auxiliary byte head. In contrast to logit-centric methods, we study offline quantized hidden-state trajectories on a UTF-8 byte clock at $0.00\text{ GiB}$ teacher VRAM and compare directly against offline \texttt{ULD} and causal prefix-conditioned \texttt{BLD}.

\textbf{Tokenizer-Free Byte Modeling.} Byte- and patch-level language models such as \textbf{MegaByte} \cite{yu2023megabyte} and \textbf{BLT} \cite{pagnoni2024blt} remove subword vocabularies altogether. Our \texttt{byte260} suite evaluates subword-to-byte hidden transfer where shared token-boundary overlap drops from $\sim 91.7\%$ to $\sim 19.8\%$.

\section{Method: UTF-8 Byte-Clock Offline Hidden Distillation}

\subsection{Shared UTF-8 Byte Clock and Offline Quantized Shards}
Let raw UTF-8 text $x \in \{0,\dots,255\}^L$ be tokenized by teacher $k \in \{1,\dots,K\}$ into $N_{T,k}$ tokens with strictly increasing cumulative byte endpoints $0 = b^{(k)}_0 < b^{(k)}_1 < \dots < b^{(k)}_{N_{T,k}} = L$, and by the student into $N_S$ tokens with cumulative byte endpoints $0 = s_0 < s_1 < \dots < s_{N_S} = L$. All tokenizer adapters enforce byte conservation ($b^{(k)}_{N_{T,k}} = s_{N_S} = L$) across ByteLevel BPE (\texttt{GPT-2}, \texttt{Llama-3}, \texttt{Qwen2}), SentencePiece+ByteFallback (\texttt{Gemma-4}), and raw UTF-8 bytes (\texttt{byte260}).

During a single offline sweep per teacher, we record hidden states at $D=3$ proportional depths ($50\%, 75\%, 100\%$ of layers), fit corpus-level top-$256$ PCA bases per depth, and serialize quantized trajectories alongside static per-depth training-corpus centroids $\mu_{k,d}$. Including $2\text{ B}$ \texttt{uint16} cumulative byte endpoints (which accommodate chunks up to $65{,}535\text{ B}$, or $\sim 13\text{K}$ English tokens; longer contexts can store per-token \texttt{uint8} byte-span deltas at $109\text{ B/tok}$ for tokens $\le 255\text{ B}$), we evaluate four \texttt{WF20} storage tiers across $D=3$ depths (exact packed codec footprints per teacher token):
\begin{itemize}
    \item \texttt{INT8-256}: $3 \times (256 \times 1\text{ B} + 2\text{ B scale}) + 2\text{ B} = \mathbf{776\text{ B/tok/teacher}}$.
    \item \texttt{INT4-128} (group-$32$ packed nibbles): $3 \times (64\text{ B} + 4 \times 2\text{ B}) + 2\text{ B} = \mathbf{218\text{ B/tok}}$ ($3.56\times$ smaller).
    \item \texttt{INT8-64} (rank-64 ablation, no 4-bit quantization): $3 \times (64\text{ B} + 2\text{ B}) + 2\text{ B} = \mathbf{200\text{ B/tok}}$ ($3.88\times$ smaller).
    \item \texttt{INT4-64} (group-$32$ packed nibbles): $3 \times (32\text{ B} + 2 \times 2\text{ B}) + 2\text{ B} = \mathbf{110\text{ B/tok}}$ ($7.05\times$ smaller than \texttt{INT8-256}, $1.76\times$ smaller than uncompressed Top-$32$ sparse logits at $194\text{ B/tok}$, and $4{,}766\times$ smaller than dense FP16 $262\text{K}$ logits). Note that a bit-packed Top-$32$ logit cache using $18$-bit vocabulary IDs and $8$-bit quantized log-probabilities would occupy $\lceil 32 \times 26 / 8 \rceil + 2 = 106\text{ B/tok}$---comparable to \texttt{INT4-64} in disk footprint, though still supplying only next-token logits rather than multi-depth representations.
\end{itemize}
To ablate all four tiers cleanly from one teacher sweep without re-sweeping disk shards per tier, our training runs memory-map the uncentered \texttt{INT8-256} research shards on disk ($776\text{ B/tok}$ wave payload; $\approx 877$--$878\text{ B/tok}$ including auxiliary Mode-B interference arrays on GPT-2), simulate group-$32$ \texttt{INT4} truncation on the fly (\texttt{simulate\_int4\_truncation\_from\_int8}), and subtract $\mu_{k,d}$ afterward. Pre-training check \texttt{gate0e} confirms $0.0$ roundtrip error between packed bit-level \texttt{INT4} shards and on-the-fly simulation; quantizing prior to centroid subtraction is conservative for accuracy (preserving the DC offset inside the 4-bit range) while adding minor simulation compute overhead during training.
For a student of width $d_S=640$, each captured depth $d \in \{6, 9, 12\}$ projects its residual stream $h_S^{(d)} \in \mathbb{R}^{640}$ through bias-free linear probes $P_{k,d} \in \mathbb{R}^{r \times d_S}$ discarded at inference. Across $K=3$ teachers, the combined per-depth probe rank $K \cdot r$ is over-determined at $r=256$ ($768 > 640$) and well-conditioned at $r=64$ ($192 < 640$, $30\%$ of residual capacity).

\subsection{Discrete Boundary Alignment (\texttt{StepAlign}, \texorpdfstring{$H=0$}{H=0})}
For student position $j \ge 5$ (skipping the first 4 warmup tokens), let $\mathcal{M}_k(j) = \mathbb{I}[\exists\, i : b^{(k)}_i = s_j]$ indicate a coincident UTF-8 byte boundary with teacher index $i(j)$. Letting $h_{k,d}[i] \in \mathbb{R}^r$ denote the raw projected teacher state and $\tilde{h}_{k,d}[i] = \mathrm{norm}(h_{k,d}[i] - \mu_{k,d})$ its centered, unit-normalized target, with student projection $p_{S,k,d}[j] = \mathrm{norm}(P_{k,d} h_S^{(d)}[j])$, \texttt{StepAlign} minimizes:
\begin{equation}
\mathcal{L}_{H=0}^{(k)} = \frac{1}{D} \sum_{d=1}^{D} \frac{\sum_{j=5}^{N_S} \mathcal{M}_k(j)\,\big(1 - \langle p_{S,k,d}[j],\, \tilde{h}_{k,d}[i(j)]\rangle\big)}{\max\big(1,\, \sum_{j=5}^{N_S} \mathcal{M}_k(j)\big)}.
\label{eq:stepalign}
\end{equation}
Equation~\eqref{eq:stepalign} is causal by construction ($b^{(k)}_i = s_j$), requires only a vectorized \texttt{searchsorted} lookup without custom CUDA kernels, and extends to $K$ teachers via $\mathcal{L}_{\mathrm{aux}} = \sum_{k=1}^K \alpha_k \mathcal{L}_{H=0}^{(k)}$ ($\sum_k \alpha_k = 1$). Across all distillation methods (\texttt{StepAlign}, \texttt{WaveDistill}, \texttt{ULD}, and \texttt{BLD}), the total training objective is strictly additive:
\begin{equation}
\mathcal{L}_{\mathrm{total}} = \mathcal{L}_{\mathrm{CE}} + w\,\mathcal{L}_{\mathrm{aux}},
\label{eq:total_loss}
\end{equation}
preserving full weight on next-token cross-entropy rather than interpolating $(1-w)\mathcal{L}_{\mathrm{CE}} + w\,\mathcal{L}_{\mathrm{aux}}$.

\subsection{Continuous-Time Causal Wave Resampling (\texttt{WaveDistill}, \texorpdfstring{\mbox{$H > 0$}}{H > 0})}
\label{sec:dc_cone}
To test whether dense supervision across non-coincident boundaries ($\mathcal{M}_k(j)=0$) improves transfer, we represent a projected teacher trajectory as a piecewise-constant signal $x(u) = x_i$ on $(b_{i-1}, b_i]$ (taking $x_i = h_i$ when uncentered and $x_i = \tilde{h}_i = \mathrm{norm}(h_i - \mu_{k,d})$ when centered).

\begin{definition}[Causal Constant-$Q$ Byte-Clock Laplace--Fourier Wave Operator]
For $M$ logarithmically spaced byte horizons $H_m \in [H_{\min}, H_{\max}]$, decay rates $\gamma_m = 1/H_m$, and frequencies $\omega_m = q/H_m$ ($q=0$: real Laplace decay; $q=1.25$: complex Laplace--Fourier wave), let $\bar{\imath}(j) = \max\{i : b_i \le s_j\}$ be the index of the last teacher token completed at or before student byte boundary $s_j$ (with $b_0 = 0$). The strictly causal normalized wave state at $s_j$ is:
\begin{equation}
W_m(s_j) = \frac{1}{Z_m(s_j)} \int_{0}^{b_{\bar{\imath}(j)}} e^{(-\gamma_m + i\omega_m)(s_j - u)}\, x(u)\, du, \qquad Z_m(s_j) = \int_{0}^{b_{\bar{\imath}(j)}} e^{-\gamma_m (s_j - u)}\, du.
\label{eq:wave_op}
\end{equation}
Because both the forcing integral and normalizer $Z_m(s_j)$ integrate only up to $b_{\bar{\imath}(j)} \le s_j$, any teacher token straddling $s_j$ ($b_{\bar{\imath}(j)} < s_j < b_{\bar{\imath}(j)+1}$) contributes zero forcing and the state evolves over $(b_{\bar{\imath}(j)}, s_j]$ purely by homogeneous decay $e^{(-\gamma_m + i\omega_m)(s_j - b_{\bar{\imath}(j)})}$ (with $Z_m$ decaying by $e^{-\gamma_m (s_j - b_{\bar{\imath}(j)})}$), guaranteeing zero future-byte leakage.
\end{definition}

Integrating Eq.~\eqref{eq:wave_op} over each completed span $\Delta b_i = b_i - b_{i-1}$ yields the $O(N_T + N_S)$ two-pointer recurrence in Eq.~\eqref{eq:recurrence}, implemented in an NVRTC CUDA kernel (\texttt{wd\_causal\_wave\_fwd\_k}) together with its reverse-time complex adjoint backward pass (\texttt{wd\_self\_wave\_bwd\_k}):
\begin{equation}
a_m \leftarrow e^{(-\gamma_m + i\omega_m)\Delta b_i} a_m + \frac{1 - e^{(-\gamma_m + i\omega_m)\Delta b_i}}{\gamma_m - i\omega_m}\, x_i.
\label{eq:recurrence}
\end{equation}
To supervise the student at every byte boundary $s_j$, we normalize the teacher's resampled complex wave in $\mathbb{C}^r$ as $\hat{W}_m^{T,(k,d)}(s_j) = W_m^{T,(k,d)}(s_j) / \|W_m^{T,(k,d)}(s_j)\|_2$, pass the unit-normalized student projection $p_{S,k,d}[j] = \mathrm{norm}(P_{k,d} h_S^{(d)}[j])$ through the same causal wave operator on the student's own spans $(s_{j-1}, s_j]$ (differentiated via \texttt{wd\_self\_wave\_bwd\_k}) to obtain $\hat{W}_m^{S,(k,d)}(s_j) \in \mathbb{C}^r$, and minimize the complex cosine loss:
\begin{equation}
\mathcal{L}_{\mathrm{geo}}^{(k)} = \frac{1}{D M (N_S - 4)} \sum_{d=1}^{D} \sum_{m=1}^{M} \sum_{j=5}^{N_S} \Big(1 - \mathrm{Re}\,\big\langle \hat{W}_m^{S,(k,d)}(s_j),\, \hat{W}_m^{T,(k,d)}(s_j)\big\rangle\Big).
\label{eq:wave_geo}
\end{equation}

\begin{corollary}[Intra-Word Phase Degeneracy at $q=0$ and Causal Information Bottleneck]
\label{cor:intraword}
For any student byte $s_j \in (b_{\bar{\imath}(j)}, b_{\bar{\imath}(j)+1})$ straddling an incomplete multi-byte teacher token with $\bar{\imath}(j) \ge 1$ and $\|W_m^T(b_{\bar{\imath}(j)})\|_2 > 0$, Eq.~\eqref{eq:wave_op} followed by $\mathbb{C}^r$ unit normalization yields the exact identity:
\begin{equation}
\hat{W}_m^{T}(s_j) = e^{i\omega_m (s_j - b_{\bar{\imath}(j)})}\, \hat{W}_m^{T}(b_{\bar{\imath}(j)}).
\label{eq:intraword_phase}
\end{equation}
\end{corollary}
\begin{proof}
Because no teacher token completes in $(b_{\bar{\imath}(j)}, s_j]$ and $\bar{\imath}(j) \ge 1$ ensures $Z_m(b_{\bar{\imath}(j)}) > 0$, the forcing integral in Eq.~\eqref{eq:wave_op} scales by $e^{(-\gamma_m + i\omega_m)(s_j - b_{\bar{\imath}(j)})}$ while $Z_m(s_j)$ scales by $e^{-\gamma_m (s_j - b_{\bar{\imath}(j)})}$. The real decay factor cancels identically in $W_m^T(s_j)$ (and in $\|W_m^T(s_j)\|_2 > 0$), leaving only the unit-modulus phase rotation $e^{i\omega_m (s_j - b_{\bar{\imath}(j)})}$.
\end{proof}

Corollary~\ref{cor:intraword} establishes two structural limits of offline causal wave supervision. First, when $q=0$ ($\omega_m = 0$), $\hat{W}_m^T(s_j) \equiv \hat{W}_m^T(b_{\bar{\imath}(j)})$ collapses to a piecewise-constant target across all intra-word bytes while the student's own wave $\hat{W}_m^S(s_j)$ integrates changing student byte states---directly explaining our empirical regime-dependent phase effect on \texttt{byte260} (Section~\ref{sec:byte260}). Second, any offline teacher scheme faces a fundamental causal information bottleneck on $(b_{\bar{\imath}(j)}, b_{\bar{\imath}(j)+1})$: interpolating toward the future teacher state at $b_{\bar{\imath}(j)+1}$ would leak unseen future bytes of the straddled word into causal student states unavailable at inference, whereas any strictly causal filter adds zero new teacher information between completed subword boundaries while temporally blurring sharp targets at coincident boundaries. Additionally, while teacher inputs $\tilde{h}_{k,d}[i]$ are corpus-centered before wave integration, $p_{S,k,d}[j]$ is unit-normalized without centering before student wave integration, allowing a directional bias in $P_{k,d} h_S^{(d)}$ to re-introduce a student-side mean cone inside $W_m^{S,(k,d)}(s_j)$.

Our initial \texttt{WaveDistill-v1} prototype integrated uncentered states $x_i = h_i$ over $M=16$ horizons $H_m \in [4, 4096]\text{ B}$ plus an 8-lag autocorrelation loss (which reduced throughput $4\times$ to $10$--$12\text{K tok/s}$). We next analyze why uncentered exponential integration collapses semantic contrast as $H_m$ grows:

\begin{proposition}[DC Mean-Cone Dominance in Uncentered Wave Integration]
\label{prop:dc_cone}
Let an uncentered projected hidden trajectory be modeled as $h(u) = \mu + \delta(u)$ on $\mathbb{R}^r$, where $\mu = \mathbb{E}[h(u)] \neq 0$ is the static mean cone and $\delta(u)$ is a zero-mean Ornstein--Uhlenbeck process with $\mathbb{E}[\delta(u)^\top \delta(v)] = \sigma^2 e^{-|u-v|/\tau_c}$ ($\tau_c > 0$). Under steady-state integration ($s_j \gg H_m$, $b_{\bar{\imath}(j)} = s_j$, $Z_m \to 1/\gamma_m$), the squared norms of the DC and semantic components are:
\begin{equation}
\|W_m^{\mathrm{DC}}\|^2 = \frac{\|\mu\|^2}{1 + q^2}, \qquad
\mathbb{E}\!\left[\|W_m^{\mathrm{sem}}\|^2\right] = \sigma^2 \frac{\gamma_m(\gamma_m + \tau_c^{-1})}{(\gamma_m + \tau_c^{-1})^2 + \omega_m^2}.
\label{eq:exact_ou}
\end{equation}
The steady-state Signal-to-DC-Cone Ratio (SDCR) therefore equals:
\begin{equation}
\begin{aligned}
\mathrm{SDCR}(H_m)
&= \frac{\sqrt{\mathbb{E}[\|W_m^{\mathrm{sem}}\|^2]}}{\|W_m^{\mathrm{DC}}\|}
 = \frac{\sigma}{\|\mu\|}\sqrt{\frac{(1+q^2)\,\frac{\tau_c}{H_m}\left(1 + \frac{\tau_c}{H_m}\right)}{\left(1 + \frac{\tau_c}{H_m}\right)^2 + q^2\left(\frac{\tau_c}{H_m}\right)^2}} \\
&\xrightarrow{\;H_m \gg \tau_c\;}\; \frac{\sigma}{\|\mu\|}\sqrt{(1+q^2)\frac{\tau_c}{H_m}} = O\!\left(H_m^{-1/2}\right).
\end{aligned}
\label{eq:sdcr}
\end{equation}
\end{proposition}

\begin{proof}
For the constant term $\mu$, $W_m^{\mathrm{DC}} = \gamma_m \mu \int_0^\infty e^{(-\gamma_m + i\omega_m)\tau} d\tau = \frac{\gamma_m}{\gamma_m - i\omega_m}\mu$, giving squared norm $\frac{\|\mu\|^2}{1+q^2}$ since $\omega_m = q\gamma_m$. For the zero-mean fluctuation $\delta$, taking expectations yields
\[
\mathbb{E}\!\left[\|W_m^{\mathrm{sem}}\|^2\right] = \gamma_m^2 \sigma^2 \int_0^\infty\!\!\int_0^\infty e^{-\gamma_m(\tau_1+\tau_2)} \cos\!\big(\omega_m(\tau_1-\tau_2)\big)\, e^{-|\tau_1-\tau_2|/\tau_c}\, d\tau_1\, d\tau_2.
\]
Changing variables to $u = \tau_1 + \tau_2$ and $v = \tau_1 - \tau_2$ (Jacobian $\frac{1}{2}$, $u \ge |v|$) separates $\int_{|v|}^\infty e^{-\gamma_m u} du = \gamma_m^{-1} e^{-\gamma_m |v|}$, leaving
\[
\mathbb{E}\!\left[\|W_m^{\mathrm{sem}}\|^2\right] = \gamma_m \sigma^2 \int_0^\infty e^{-(\gamma_m + \tau_c^{-1})v}\cos(\omega_m v)\, dv = \sigma^2 \frac{\gamma_m(\gamma_m + \tau_c^{-1})}{(\gamma_m + \tau_c^{-1})^2 + \omega_m^2}.
\]
Dividing the semantic root-mean-square in Eq.~\eqref{eq:exact_ou} by $\|W_m^{\mathrm{DC}}\|$ and substituting $\gamma_m = 1/H_m$, $\omega_m = q/H_m$ gives Eq.~\eqref{eq:sdcr}.
\end{proof}

\emph{Scope remark:} Proposition~\ref{prop:dc_cone} is a standard steady-state EMA variance calculation ($s_j \gg H_m$). On 512-token GPT-2 training sequences (mean length $2{,}507\text{ B}$, $4.90\text{ B/tok}$; $2{,}332\text{ B}$ on the 120-seq val split), horizons $H_m \in [4, 64]\text{ B}$ (used in \texttt{WaveDistill-v2}) reach $95\%$ of steady state ($\approx 3 H_m$) within $3$--$40$ tokens, $H_m = 256\text{ B}$ requires $\sim 160$ tokens, and the largest \texttt{WaveDistill-v1} horizons ($H_m \in [1024, 4096]\text{ B}$) operate in the finite-length transient regime ($s_j \lesssim H_m$), where the finite-$s_j$ normalization factor still preserves DC accumulation over the sequence while averaging out short-range semantic variance.

In \texttt{WaveDistill-v2}, we subtract the static corpus centroid $\mu_{k,d}$ before wave integration (driving $\|W_m^{\mathrm{DC}}\| \equiv 0$ without future leakage), restrict to $M=4$ short horizons $H_m \in [4, 64]\text{ B}$, and remove the autocorrelation term ($\lambda_{\mathrm{int}}=0$), testing both pure wave ($\lambda_{\mathrm{geo}}=0.60, \lambda_{\mathrm{step}}=0$) and hybrid ($\lambda_{\mathrm{geo}}=0.25, \lambda_{\mathrm{step}}=0.35$) configurations at total auxiliary weight $w=0.60$.

\section{Experimental Protocol}

\textbf{Students and Optimization.} All arms train the same Llama-style causal Transformer ($12\text{L}$, $d_S=640$, $10\text{Q}/5\text{KV}$ GQA \cite{ainslie2023gqa}, $d_{\mathrm{ff}}=1728$ SwiGLU, RoPE \cite{su2021roformer}, RMSNorm \cite{zhang2019rmsnorm}, weight-tied embeddings) from random initialization: $86.74\text{M}$ parameters for GPT-2 ($V=50{,}257$; \cite{radford2019gpt2}) and $54.74\text{M}$ parameters for pure UTF-8 bytes (\texttt{byte260}, $V=260$). We refer to the plain next-token cross-entropy baseline trained from scratch as \texttt{CE-only / SFT} (matching codebase arm \texttt{sft}). Within each seed, every arm starts from the identical saved initialization checkpoint and processes the identical minibatch sequence order on one RTX 5080 ($16\text{ GiB}$). All 73 primary runs and offline controls run strictly at $0.00\text{ GiB}$ teacher VRAM verified prior to arm load (whereas the compact frozen $86.7\text{M}$ random-init and \texttt{GPT-2-124M} teacher-scale controls in \texttt{run\_controls.py} evaluate their frozen teacher online for convenience). Training uses effective batch size $32 \times 512 = 16{,}384$ student tokens/step ($640$ steps for 5 epochs; $128$ steps for 1 epoch), \texttt{bf16} autocast, gradient-norm clipping at $1.0$, fused AdamW ($\beta_1=0.9, \beta_2=0.95$, $\text{lr}=6\times 10^{-4}\to 6\times 10^{-5}$ cosine after linear warmup of $\max(20, \lfloor 0.06 \cdot T\rfloor)$ steps, $\text{wd}=0.10$ on 2D weights, $\text{wd}=0.0$ on norms/biases, dropout $=0.0$), and $1.5\times$ LR with $\text{wd}=0.0$ on auxiliary distillation probes/heads. Regularized CE-only controls additionally sweep dropout ($0.10$) and weight decay ($0.20, 0.30$).

\textbf{Teachers.} Three open-weight models in Ollama \texttt{Q4\_K\_M} GGUF format (\texttt{file\_type=15}; \texttt{Q4\_K}/\allowbreak\texttt{Q6\_K} weights, \texttt{F32} norms) are swept sequentially via layer streaming and unloaded ($<0.02\text{ GiB}$ residual VRAM): (i)~\texttt{Llama-3.2-\allowbreak 3B-Instruct} ($2.02\text{ GB}$, $28\text{L}$, $d=3072$, $V=128{,}256$, depths $14/21/28$); (ii)~\texttt{Gemma-4-E4B} ($5.49\text{ GB}$, $7.5\text{B}$ total / $4\text{B}$ effective active, $42\text{L}$, $d=2560$, $V=262{,}144$, depths $21/32/42$; \cite{gemmateam2026gemma4}); and (iii)~\texttt{DeepSeek-\allowbreak R1-\allowbreak Distill-\allowbreak Qwen-14B} ($8.99\text{ GB}$, $48\text{L}$, $d=5120$, $V=152{,}064$, depths $24/36/48$; \cite{deepseek2025r1}). As a 3-seed small pretrained teacher-scale control, we additionally evaluate pretrained \texttt{GPT-2-124M} ($12\text{L}, d=768$, $V=50{,}257$, depths $6/9/12$, corpus top-$64$ PCA capturing $97.4\%/94.6\%/99.4\%$ of variance + group-$32$ \texttt{INT4} + DC-centering, masked at the identical \texttt{Llama-3.2} boundary positions across all three seeds \texttt{20261126}--\texttt{20261128}).

\textbf{Corpus, Regimes, and Evaluation.} Training bundles ($4{,}096 \times 512 = 2{,}097{,}152$ unique student tokens) mix Cosmopedia-100k shard 0, SciQ-train, ARC-Challenge-train, SQuAD\_v2-train, and Python-codes-25k (\texttt{[:16000]}), hash-filtered for $0\%$ overlap against evaluation splits. Because both suites match student token count ($2.10\text{M}$ tokens), the GPT-2 bundle spans $10.27\text{M}$ unique UTF-8 bytes ($4.90\text{ B/tok}$) whereas the \texttt{byte260} bundle spans $2.10\text{M}$ unique UTF-8 bytes ($1.00\text{ B/tok}$, $4.90\times$ less unique raw text). Crucially, the same span difference holds during evaluation: $120 \times 512$ tokens cover $279{,}867$ UTF-8 bytes on the GPT-2 validation split ($276{,}581\text{ B}$ on fresh test) vs.\ $61{,}440$ UTF-8 bytes on \texttt{byte260} (the leading $512$-byte prefixes of each packed sequence). Consequently, BPB metrics are strictly comparable \emph{within} each tokenizer suite, never directly across GPT-2 vs.\ \texttt{byte260} ($1.587$ vs.\ $1.995$).

We execute \textbf{98 controlled runs}: (A)~5-epoch GPT-2 ($10.49\text{M}$ tokens/arm: 42 seed-1 runs covering all baselines, teachers, and ablations, plus 6 core arms across seeds 2 and 3, \texttt{20261126}--\texttt{20261128}); (B)~1-epoch zero-repeat GPT-2 (8 primary arms + 2 null/untrained controls, $2.10\text{M}$ unique tokens); (C)~5-epoch pure UTF-8 \texttt{byte260} (11 primary arms + 4 \texttt{Byte BLD} weight controls $w \in \{0.10, 0.20, 1.00, 2.00\}$, $10.49\text{M}$ bytes/arm, single seed \texttt{20261126}); and (D)~19 additional 5-epoch GPT-2 controls (frozen random-init Transformer teachers with PCA+\texttt{INT4} and orthogonal projections, 3-seed pretrained \texttt{GPT-2-124M} teacher matching the exact saved init and batch order of seeds 1--3, shuffled-document, shuffled-position, and batch-seeded random-Gaussian targets; regularized CE-only/SFT; extended \texttt{BLD} $w \in \{0.05, 0.02\}$; exact offline \texttt{ULD} at $\tau=1.0, w \in \{0.10, 0.50\}$; within-batch \texttt{INT8-256} and \texttt{INT8-64} rank-vs-bitwidth controls; and full-120-seq per-epoch curves). Primary metrics are unseen multi-domain held-out Perplexity and Bits-Per-Byte (BPB) across $120$ validation sequences ($40$ each from Cosmopedia shard-1 tail, SciQ-val, and Python tail \texttt{[23613:]}). For every paired comparison over the $120$ sequences, we report \emph{both} $B=100{,}000$ nonparametric paired bootstrap resamples (two-sided $2k/B$; zero sign crossings out of $100{,}000$ lie below Monte Carlo resolution $2\times 10^{-5}$ and are reported as $<2\times 10^{-5}$ with exact crossing count $(0)$) \emph{and} exact analytic two-sided paired $t$-tests ($\mathrm{df}=119$, $p_t$, resolving arbitrarily small tail probabilities without Monte Carlo floor; per-sequence NLLs for all 25 controls are saved in \path{results/controls_benchmark.json}), plus exact two-sided Welch's $t$-tests on 3-seed means.

\textbf{Independent Fresh Test Split (Disclosing Initial Split Revision).} To verify results free of validation-set reuse, we first built a 64-byte-filtered fresh split using Cosmopedia shard-1 \texttt{[0:800]}, SciQ-\texttt{test}, and Python \texttt{[17000:20000]}. Subsequent inspection revealed that \texttt{[17000:20000]} sat adjacent to training \texttt{[:16000]} with $9.41\%$ raw 32-byte shingle overlap and much lower code entropy ($1.41$ vs.\ $2.22$ BPB; on that initial split, 3-seed \texttt{StepAlign 1T} beat SFT by $\Delta\text{BPB}=-0.03250$, BPB Welch $p=2.79\times 10^{-5}$, while \texttt{StepAlign 1T} vs.\ \texttt{Wave-v2 Hybrid 1T} was non-significant at Welch $p=0.271$). To eliminate near-duplicate drift, we replaced the Python slice with the tail-adjacent \texttt{[22000:23600]} regime (matching val \texttt{[23613:]} with $0.06\%$ raw overlap), tightened cryptographic deduplication to strictly $0$ 32-byte overlaps against both train and val, and re-scored all \textbf{37} saved checkpoints without omission (\path{results/fresh_test_split_audit.json}, Table~\ref{tab:fresh_test}). Zero-shot 4-choice accuracy on \texttt{SciQ-test} ($1{,}000$), \texttt{ARC-Challenge-test} ($1{,}172$), and \texttt{HellaSwag-val} ($1{,}000$) is reported for completeness; at $55\text{M}$--$87\text{M}$ parameters these remain near the $25\%$ chance floor (SE $\pm 1.4\%$).

\textbf{Baselines and Pre-Training Verification.} All baselines run offline at $0.00\text{ GiB}$ teacher VRAM from Top-$32$ sparse cached teacher logits ($32 \times (4\text{ B int32} + 2\text{ B fp16}) + 2\text{ B uint16} = 194\text{ B/tok}$): plain and regularized \texttt{CE-only / SFT}; \texttt{ULD} \cite{boizard2025uld} across six configurations ($w=1.0, \tau=1.5$ all-causal; $w \in \{0.05, 0.10, 0.25\}, \tau=2.0$ exact-boundary; and exact $\tau=1.0, w \in \{0.10, 0.50\}$ exact-boundary); and \texttt{BLD} / \texttt{BLD-3T} \cite{singh2026bld} (causal prefix-conditioned next-byte marginalization over Top-$32$ log-probs supervising a dedicated discarded auxiliary linear head $\mathbb{R}^{d_S} \to \mathbb{R}^{256}$ on both GPT-2 across $w \in \{0.02, 0.05, 0.10, 0.20, 0.50\}$ and \texttt{byte260} across $w \in \{0.10, 0.20, 0.50, 1.00, 2.00\}$). Note that for offline Top-$32$ \texttt{ULD} at $\tau > 1$, tempering the aggregated residual tail mass as $(1 - \sum_{i \le 32} p_i)^{1/\tau}$ introduces a Jensen's-inequality mismatch relative to full-vocabulary softmax tempering, whereas $\tau = 1.0$ is exact; we therefore include both regimes in Table~\ref{tab:controls}. Pre-training verification (\path{results/gate0_verify_report.json}) confirmed across \emph{all three teachers}: ``\texttt{The capital of France is}'' $\to$ ``\texttt{ Paris}'' top-1 sanity checks ($p=0.6155$ Llama, $0.9952$ Gemma, $0.7967$ DeepSeek) and exact online chunked Top-$32$ \texttt{log\_softmax} parity against full-vocabulary softmax---top-1 exact and $32/32$ tie-exact Top-$32$ membership for all three (\texttt{Llama-3.2-3B}: $32/32$ strict overlap, max diff $8.89 \times 10^{-4}$, $0.0089\text{ GiB}$ post-unload; \texttt{Gemma-4-E4B}: $31/32$ strict due to a 4-way exact \texttt{bf16} tie at rank 32, max diff $2.70 \times 10^{-3}$, $0.0108\text{ GiB}$; \texttt{DeepSeek-R1-Qwen-14B}: $31/32$ strict due to a 2-way exact \texttt{bf16} tie at rank 32, max diff $1.46 \times 10^{-3}$, $0.0110\text{ GiB}$); zero causality leakage across all $3 \times 2$ teacher--student pairs on straddled multi-byte tokens; subword-split wave invariance ($\min\cos(H \ge 16\text{B}) = 0.99498$); CUDA forward parity ($2.4 \times 10^{-7}$) and adjoint backward gradcheck ($3.81 \times 10^{-6}$ full, $2.38 \times 10^{-6}$ real, $4.8 \times 10^{-7}$ $M=1$); and $0.0$ roundtrip error between packed and simulated \texttt{INT4} codecs.

\begin{table}[t]
\centering
\caption{Aligned bundle spectral geometry and UTF-8 boundary overlap across GPT-2 ($50\text{K}$) and Byte-260 ($260$) in \texttt{paper\_statistical\_audit.json}. Boundary overlap is measured over the first $512$ training sequences; DC-cone cosines, adjacent-token cosines, and Top-64/128 variance shares (within the stored Top-256 PCA subspace) are computed over the first $64$ training sequences at the final captured depth ($100\%$ of layers). Logged full-$d_{\mathrm{model}}$ PCA-256 captured variance at the final depth (\texttt{shards\_*/*\_meta.json}) is $61.5\%$ (DeepSeek, $d=5120$) and $46.7\%$ (Gemma-4, $d=2560$) on GPT-2 ($67.3\%$, $48.8\%$, and Llama-3.2 $58.4\%$ on Byte-260), so Top-64 captures $\approx 29\%$--$51\%$ of full-$d_{\mathrm{model}}$ variance on GPT-2 ($\approx 31\%$--$57\%$ on Byte-260). On the $64$-seq slice, DeepSeek DC-cone cosines are $0.45704$ (GPT-2) and $0.45700$ (Byte-260); across all $4{,}096$ sequences they are $0.4678$ vs.\ $0.4652$.}
\label{tab:diagnostics}
\resizebox{\textwidth}{!}{%
\begin{tabular}{llcccc}
\toprule
Student & Teacher & Boundary Overlap & DC Cone $\cos$ & Adj.\ $\cos$ (Raw $\to$ Cen.) & Top-64 / 128 of PCA-256 \\
\midrule
\multirow{3}{*}{GPT-2 ($50\text{K}$)}
  & \texttt{Llama-3.2-3B} ($128\text{K}$)           & $91.35\%$ & $0.5711$ & $0.6044 \to 0.4229$ & $69.8\%$ / $83.1\%$ \\
  & \texttt{Gemma-4-E4B} ($262\text{K}$)            & $92.34\%$ & $0.3229$ & $0.4018 \to 0.3330$ & $61.6\%$ / $78.5\%$ \\
  & \texttt{DeepSeek-R1-Qwen-14B} ($152\text{K}$)   & $91.39\%$ & $0.4570$ & $0.5429 \to 0.4276$ & $82.3\%$ / $91.1\%$ \\
\multicolumn{2}{l}{\quad\textit{3-Teacher Union / Intersection}} & \textit{$95.09\%$ / $88.61\%$} & --- & --- & --- \\
\midrule
\multirow{3}{*}{Byte-260 ($260$)}
  & \texttt{Llama-3.2-3B} ($128\text{K}$)           & $19.66\%$ & $0.5940$ & $0.6109 \to 0.4088$ & $72.9\%$ / $85.0\%$ \\
  & \texttt{Gemma-4-E4B} ($262\text{K}$)            & $19.96\%$ & $0.3351$ & $0.4015 \to 0.3246$ & $63.1\%$ / $79.4\%$ \\
  & \texttt{DeepSeek-R1-Qwen-14B} ($152\text{K}$)   & $19.83\%$ & $0.4570$ & $0.5337 \to 0.4158$ & $85.0\%$ / $92.7\%$ \\
\multicolumn{2}{l}{\quad\textit{3-Teacher Union / Intersection}} & \textit{$20.89\%$ / $18.73\%$} & --- & --- & --- \\
\bottomrule
\end{tabular}%
}
\end{table}

\FloatBarrier
\section{Empirical Results}

\subsection{Three-Seed GPT-2 Comparison, Teacher-Scale, and Null / Regularized Controls}

\begin{table}[t]
\centering
\setlength{\tabcolsep}{4pt}
\caption{\textbf{3-Seed Controlled GPT-2 Core Comparison:} Mean $\pm$ std across 3 matched random seeds (\texttt{20261126}--\texttt{20261128}), $86.74\text{M}$ Transformer, $10.49\text{M}$ tokens/arm, matched total auxiliary weight $w=0.60$ and \texttt{INT4-64} ($110\text{ B/tok}$) across hidden distillation arms. All primary rows run strictly at $0.00\text{ GiB}$ teacher VRAM; row $^\dagger$ evaluates the compact frozen same-tokenizer \texttt{GPT-2-124M} teacher online in \texttt{results/controls\_benchmark.json} using the exact same saved init checkpoints and batch order for seeds 1--3. Best value per column in bold.}
\label{tab:3seed}
\resizebox{\textwidth}{!}{%
\begin{tabular}{lrrrrrrrr}
\toprule
Method (3 Seeds) & Val PPL $\downarrow$ & Val BPB $\downarrow$ & Prose BPB & Science BPB & Code BPB & Mean MC & tok/s $\uparrow$ & Peak VRAM \\
\midrule
CE-only / SFT (from scratch)              & $150.06 \pm 1.08$ & $1.58708 \pm 0.00228$ & $1.3151$ & $1.4022$ & $2.2182$ & $27.75 \pm 0.57\%$ & $\mathbf{63.1\text{K}}$ & $\mathbf{5.62\text{G}}$ \\
ULD (Llama-3.2, $w=0.05$, boundary)       & $150.62 \pm 1.60$ & $1.58827 \pm 0.00335$ & $1.3156$ & $1.4031$ & $2.2207$ & $27.59 \pm 0.27\%$ & $41.1\text{K}$ & $7.92\text{G}$ \\
StepAlign$^\dagger$ (\texttt{GPT-2-124M}, PCA-64+\texttt{INT4}) & $137.11 \pm 0.66$ & $1.55851 \pm 0.00152$ & $1.2887$ & $1.3705$ & $2.1904$ & --- & $44.2\text{K}$ & online \\
\textbf{StepAlign (Llama-3.2, INT4-64, $w=0.60$)} & $\mathbf{134.76 \pm 0.40}$ & $\mathbf{1.55303 \pm 0.00095}$ & $\mathbf{1.2828}$ & $\mathbf{1.3695}$ & $2.1799$ & $\mathbf{28.09 \pm 0.54\%}$ & $49.9\text{K}$ & $5.67\text{G}$ \\
Wave-v2 Hybrid (Llama-3.2, $0.25/0.35$)   & $136.15 \pm 0.57$ & $1.55627 \pm 0.00133$ & $1.2852$ & $1.3703$ & $2.1873$ & $27.95 \pm 0.17\%$ & $44.6\text{K}$ & $5.66\text{G}$ \\
StepAlign 3-Teacher (INT4-64, $w=0.60$)   & $136.26 \pm 0.64$ & $1.55653 \pm 0.00150$ & $1.2881$ & $1.3742$ & $\mathbf{2.1791}$ & $27.84 \pm 0.29\%$ & $37.3\text{K}$ & $5.67\text{G}$ \\
Wave-v2 Pure 3-Teacher (INT4-64, $w=0.60$)& $137.97 \pm 0.94$ & $1.56048 \pm 0.00215$ & $1.2892$ & $1.3748$ & $2.1915$ & $27.98 \pm 0.49\%$ & $34.4\text{K}$ & $5.67\text{G}$ \\
\bottomrule
\end{tabular}%
}
\end{table}

\begin{table}[t]
\centering
\setlength{\tabcolsep}{4pt}
\caption{\textbf{Controlled Ablation Suite: Null/Untrained Teachers, 3-Seed \texttt{GPT-2-124M}, Regularized CE-Only, Extended BLD, Exact Offline ULD ($\tau=1.0$), and Within-Batch Rank-vs-Bitwidth Factorization} (GPT-2 $86.74\text{M}$, identical saved init and batch order per seed, full-120-seq per-epoch evaluation). Per-sequence NLLs for all 25 control runs (19 shown here, 2 one-epoch controls in Table~\ref{tab:1ep_and_teachers}, and 4 \texttt{byte260} \texttt{BLD} controls in Table~\ref{tab:byte260}) are saved in \texttt{results/controls\_benchmark.json}. Best value per column in bold.}
\label{tab:controls}
\resizebox{\textwidth}{!}{%
\begin{tabular}{llrrrrrr}
\toprule
Control Group & Arm Description & Ep.\ 1 & Ep.\ 2 & Ep.\ 3 & Ep.\ 4 & Final (Ep.\ 5) $\downarrow$ & Val BPB $\downarrow$ \\
\midrule
\multirow{5}{*}{\shortstack[l]{Null \& Untrained\\Teacher Controls\\(Seed 1, $d=64$,\\ $w=0.60$, boundary)}}
  & Frozen Random-Init $86.7\text{M}$ Teacher (PCA-64 + \texttt{INT4}) & $510.30$ & $285.31$ & $209.81$ & $184.04$ & $179.01$ & $1.64296$ \\
  & Frozen Random-Init $86.7\text{M}$ Teacher (Orth.\ 64-D, FP32) & $444.02$ & $255.93$ & $186.90$ & $166.47$ & $164.43$ & $1.61606$ \\
  & Cross-Document Shuffled Targets (roll batch)   & $423.28$ & $231.75$ & $170.91$ & $152.01$ & $151.51$ & $1.59013$ \\
  & Random Gaussian Targets ($d=64$, batch-seeded) & $413.97$ & $225.87$ & $168.00$ & $149.49$ & $149.14$ & $1.58514$ \\
  & Within-Seq.\ Position-Shuffled Targets         & $414.47$ & $228.39$ & $168.44$ & $148.47$ & $147.91$ & $1.58253$ \\
\midrule
\multirow{4}{*}{\shortstack[l]{CE-Only / SFT Curve\\\& Regularized SFT\\(Seed 1, Full 120-Seq)}}
  & CE-only / SFT Default ($\text{wd}=0.10$, drop $=0.0$)    & $419.29$ & $229.09$ & $167.65$ & $149.75$ & $149.21$ & $1.58529$ \\
  & SFT + Dropout ($\text{wd}=0.10$, drop $=0.10$) & $449.00$ & $248.66$ & $182.08$ & $156.38$ & $148.27$ & $1.58330$ \\
  & SFT + High WD ($\text{wd}=0.30$, drop $=0.0$)  & $407.30$ & $222.66$ & $164.41$ & $147.97$ & $146.70$ & $1.57992$ \\
  & SFT + Dropout + WD ($\text{wd}=0.20$, drop $=0.10$) & $447.78$ & $247.15$ & $178.34$ & $154.42$ & $146.59$ & $1.57969$ \\
\midrule
\multirow{4}{*}{\shortstack[l]{Extended BLD\\\& Exact Offline ULD\\(Seed 1, $\tau=1.0$)}}
  & BLD (Llama-3.2, $w=0.05$; peak $w=0.10$: $146.24$) & $414.49$ & $228.83$ & $167.96$ & $149.54$ & $149.47$ & $1.58584$ \\
  & BLD (Llama-3.2, $w=0.02$)                      & $429.62$ & $230.23$ & $169.84$ & $150.81$ & $150.10$ & $1.58717$ \\
  & ULD Exact ($\tau=1.0, w=0.10$, boundary)       & $425.36$ & $233.19$ & $171.01$ & $150.65$ & $149.56$ & $1.58604$ \\
  & ULD Exact ($\tau=1.0, w=0.50$, boundary)       & $450.54$ & $244.66$ & $177.69$ & $156.32$ & $155.91$ & $1.59919$ \\
\midrule
\multirow{3}{*}{\shortstack[l]{3-Seed Small Teacher\\(\texttt{GPT-2-124M}, $w=0.60$,\\ PCA-64 + \texttt{INT4})}}
  & Pretrained \texttt{GPT-2-124M} (seed 1: \texttt{20261126}) & $\mathbf{384.43}$ & $211.99$ & $154.00$ & $137.19$ & $136.36$ & $1.55676$ \\
  & Pretrained \texttt{GPT-2-124M} (seed 2: \texttt{20261127}) & $389.71$ & $\mathbf{208.77}$ & $155.03$ & $135.72$ & $137.54$ & $1.55950$ \\
  & Pretrained \texttt{GPT-2-124M} (seed 3: \texttt{20261128}) & $396.80$ & $212.02$ & $155.32$ & $137.74$ & $137.44$ & $1.55927$ \\
\midrule
\multirow{3}{*}{\shortstack[l]{Within-Batch Rank\\vs.\ Bit-Width\\(Seed 1, Llama-3.2, $w=0.60$)}}
  & StepAlign \texttt{INT8-256} ($776\text{ B/tok}$, full) & $392.96$ & $213.75$ & $155.92$ & $138.95$ & $137.52$ & $1.55945$ \\
  & \textbf{StepAlign \texttt{INT8-64} ($200\text{ B/tok}$, rank only)} & $389.95$ & $210.49$ & $\mathbf{153.18}$ & $\mathbf{135.11}$ & $\mathbf{134.43}$ & $\mathbf{1.55224}$ \\
  & StepAlign \texttt{INT4-64} ($110\text{ B/tok}$, rank + INT4)         & $399.29$ & $213.71$ & $155.63$ & $136.89$ & $134.77$ & $1.55305$ \\
\bottomrule
\end{tabular}%
}
\end{table}

\begin{figure}[t]
\centering
\resizebox{\textwidth}{!}{%
\begin{tikzpicture}
\begin{axis}[
    name=leftplot,
    width=8.2cm,
    height=6.8cm,
    title={\small (a) Full 120-Seq Validation PPL Across Epochs 1--5},
    xlabel={\small Training Epoch ($2.10\text{M}$ tokens/epoch)},
    ylabel={\small Validation PPL $\downarrow$},
    xmin=1, xmax=5,
    xtick={1,2,3,4,5},
    ymin=120, ymax=535,
    ymajorgrids=true,
    grid style={dashed, gray!30},
    tick label style={font=\scriptsize},
    legend style={at={(0.98,0.98)}, anchor=north east, font=\scriptsize, fill=white, fill opacity=0.92, draw=gray!50, cells={anchor=west}},
    every axis plot/.append style={line width=1.25pt, mark size=1.8pt}
]
\addplot[color=red!75!black, dashdotted, mark=triangle*] coordinates {(1,510.30)(2,285.31)(3,209.81)(4,184.04)(5,179.01)};
\addlegendentry{Rand-Init $86.7\text{M}$ (PCA+\texttt{INT4})}
\addplot[color=orange!80!black, densely dotted, mark=diamond*] coordinates {(1,423.28)(2,231.75)(3,170.91)(4,152.01)(5,151.51)};
\addlegendentry{Shuffled-Doc Control}
\addplot[color=black, dashed, mark=square*] coordinates {(1,419.29)(2,229.09)(3,167.65)(4,149.75)(5,149.21)};
\addlegendentry{Plain CE-only / SFT}
\addplot[color=gray!75!black, dashed, mark=o] coordinates {(1,447.78)(2,247.15)(3,178.34)(4,154.42)(5,146.59)};
\addlegendentry{Reg.\ SFT (drop+wd)}
\addplot[color=violet!75!black, densely dashed, mark=pentagon*] coordinates {(1,384.43)(2,211.99)(3,154.00)(4,137.19)(5,136.36)};
\addlegendentry{GPT-2-124M (PCA+\texttt{INT4})}
\addplot[color=teal!80!black, solid, mark=*] coordinates {(1,399.29)(2,213.71)(3,155.63)(4,136.89)(5,134.77)};
\addlegendentry{StepAlign \texttt{INT4-64} ($110\text{ B}$)}
\addplot[color=blue!75!black, solid, mark=+] coordinates {(1,389.95)(2,210.49)(3,153.18)(4,135.11)(5,134.43)};
\addlegendentry{StepAlign \texttt{INT8-64} ($200\text{ B}$)}
\end{axis}

\begin{axis}[
    at={(leftplot.east)}, anchor=west, xshift=3.25cm,
    width=6.8cm,
    height=6.8cm,
    title={\small (b) Converged Epoch-5 PPL ($\downarrow$)},
    xbar,
    bar shift=0pt,
    bar width=7pt,
    xmin=118, xmax=196,
    xlabel={\small Final Validation PPL $\downarrow$},
    symbolic y coords={Rand-Init PCA+4b, Rand-Init Orth, Shuffled-Doc, Plain CE/SFT, Rand-Gauss, Shuffled-Pos, Reg. SFT, BLD w=0.10, StepAlign INT8-256, GPT-2-124M 4b, StepAlign INT4-64, StepAlign INT8-64},
    ytick={Rand-Init PCA+4b, Rand-Init Orth, Shuffled-Doc, Plain CE/SFT, Rand-Gauss, Shuffled-Pos, Reg. SFT, BLD w=0.10, StepAlign INT8-256, GPT-2-124M 4b, StepAlign INT4-64, StepAlign INT8-64},
    y dir=reverse,
    enlarge y limits=0.07,
    tick label style={font=\scriptsize},
    yticklabel style={font=\scriptsize, align=right},
    xmajorgrids=true,
    grid style={dashed, gray!30},
    nodes near coords,
    nodes near coords align={horizontal},
    every node near coord/.append style={font=\scriptsize, black, xshift=1pt, fill=white, fill opacity=0.88, text opacity=1, inner sep=0.6pt},
    point meta=rawx
]
\draw[dashed, black!70, line width=0.85pt] ({axis cs:149.21,Plain CE/SFT}|-{rel axis cs:0,0}) -- ({axis cs:149.21,Plain CE/SFT}|-{rel axis cs:0,1});
\addplot[fill=red!65!white, draw=red!75!black] coordinates {(179.01,Rand-Init PCA+4b) (164.43,Rand-Init Orth)};
\addplot[fill=orange!60!white, draw=orange!80!black] coordinates {(151.51,Shuffled-Doc) (149.14,Rand-Gauss) (147.91,Shuffled-Pos)};
\addplot[fill=gray!55!white, draw=black!75] coordinates {(149.21,Plain CE/SFT) (146.59,Reg. SFT) (146.24,BLD w=0.10)};
\addplot[fill=violet!50!white, draw=violet!80!black] coordinates {(137.52,StepAlign INT8-256) (136.36,GPT-2-124M 4b)};
\addplot[fill=teal!65!white, draw=blue!70!black] coordinates {(134.77,StepAlign INT4-64) (134.43,StepAlign INT8-64)};
\end{axis}
\end{tikzpicture}%
}
\caption{\textbf{Full-120-Sequence Validation PPL Trajectories and Converged Control Separation} (GPT-2 $86.74\text{M}$, seed \texttt{20261126}). \textbf{(a)}~Plain CE-only/SFT improves monotonically through epoch~5; pretrained teachers lead from epoch~1 onward while the like-for-like PCA+\texttt{INT4} frozen random-init $86.7\text{M}$ teacher trails plain CE-only at every epoch ($510.30 \to 179.01$). \textbf{(b)}~At epoch~5, all five non-pretrained controls yield no statistically significant gain over plain CE-only/SFT (full-height dashed line, $149.21$; $p_{\text{boot}} \ge 0.079$) or actively hurt ($151.51$--$179.01$). Pretrained \texttt{GPT-2-124M} ($136.36$; 3-seed $137.11 \pm 0.66$) recovers $87\%$--$89\%$ of the seed-1 gain ($84.6\%$ across 3 seeds), while within-batch \texttt{Llama-3.2-3B} \texttt{INT8-64} ($134.43$) and \texttt{INT4-64} ($134.77$; 3-seed $134.76 \pm 0.40$) improve further over \texttt{GPT-2-124M} and by $-2.8$ to $-3.1$ PPL over within-batch \texttt{INT8-256} ($137.52$).}
\label{fig:epoch_curves}
\end{figure}
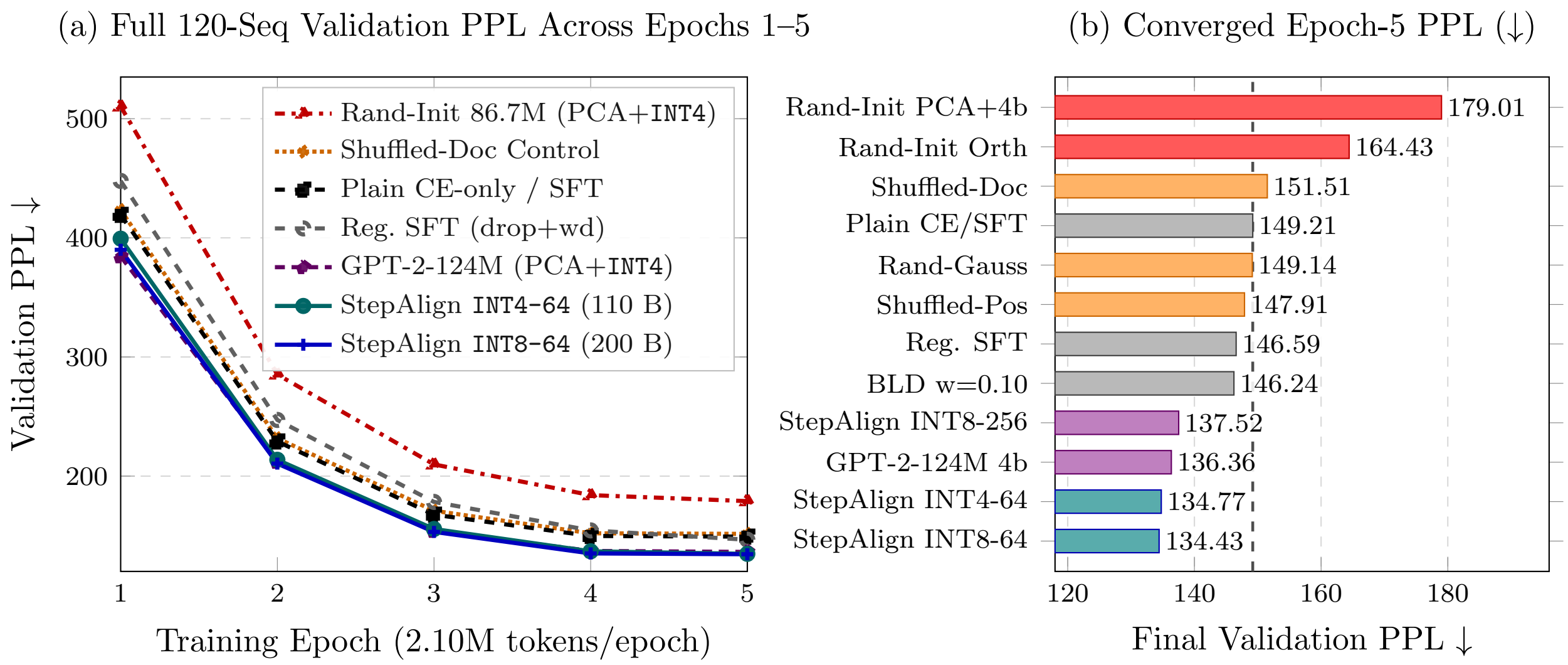

Tables~\ref{tab:3seed} and~\ref{tab:controls} and Figure~\ref{fig:epoch_curves} establish five primary conclusions:
\begin{enumerate}
    \item \textbf{Pretrained Knowledge Transfer, Generic LM Structure (3-Seed \texttt{GPT-2-124M}), and Capacity Saturation:} Because all three large teachers achieve nearly identical 5-epoch GPT-2 PPL ($134.60$--$134.82$, Table~\ref{tab:1ep_and_teachers}; $0.22$ PPL spread, well within single-seed rerun noise), two mechanistic questions arise: (a)~is the gain genuine pretrained transfer rather than probe regularization, and (b)~how much comes from generic pretrained LM structure versus a larger $3\text{B}$--$14\text{B}$ teacher? Tables~\ref{tab:3seed} and~\ref{tab:controls} and Figure~\ref{fig:epoch_curves} resolve both questions directly at identical $d=64$ probe geometry, $w=0.60$, and exact \texttt{Llama-3.2} boundary masking:
    \begin{itemize}
        \item \emph{Non-pretrained controls fail or hurt:} We evaluate a frozen randomly initialized 12-layer causal Transformer using the $86.74\text{M}$ student architecture ($12\text{L}, d=640$, GPT-2 tokens masked at \texttt{Llama-3.2} boundaries---a conservative control since it eliminates cross-tokenizer mismatch). Processed through the identical corpus top-$64$ PCA ($55.2\%$, $60.8\%$, $63.9\%$ variance across depths 6/9/12 in \texttt{results/\allowbreak controls\_\allowbreak benchmark.json}), group-$32$ \texttt{INT4}, and DC-centering pipeline, the student fits the deterministic causal targets (auxiliary loss drops $0.418 \to 0.220$), yet held-out PPL degrades to $179.01$ ($+29.80$ PPL worse than CE-only/SFT at $149.21$, $\Delta\text{BPB}=+0.05766$, $0/100\text{K}$ crossings, paired $t$-test $p_t = 2.09 \times 10^{-44}$; $+44.24$ PPL worse than real \texttt{StepAlign} at $134.77$, $p_t = 1.37 \times 10^{-67}$, $100.0\%$ win rate). A random orthogonal $640 \to 64$ FP32 projection of the same untrained teacher (aux loss $0.463 \to 0.291$) degrades PPL to $164.43$ ($+15.22$ PPL worse than SFT, $p_t = 5.96 \times 10^{-31}$). Cross-document shuffled targets degrade PPL to $151.51$ ($p_{\text{boot}}=0.00148$), batch-seeded random Gaussian targets yield $149.14$ ($p_{\text{boot}}=0.898$), and within-sequence position-shuffled targets yield $147.91$ ($p_{\text{boot}}=0.0794$, retaining document-level topic means; vs.\ real \texttt{StepAlign} $\Delta\text{BPB}=-0.02947$, $0/100\text{K}$, $p_t = 5.87 \times 10^{-34}$, $94.2\%$ win rate).
        \item \emph{Teacher scale progression ($86.7\text{M}$ untrained $\to$ 3-seed $124\text{M}$ pretrained $\to$ $3\text{B}$--$14\text{B}$):} Replacing the untrained $86.7\text{M}$ teacher with pretrained same-tokenizer \texttt{GPT-2-124M} (PCA-64 capturing $97.4\%$, $94.6\%$, $99.4\%$ of $d=768$ variance + group-$32$ \texttt{INT4}, masked at identical \texttt{Llama-3.2} boundaries) yields $136.36$, $137.54$, and $137.44$ PPL across seeds 1--3---a 3-seed mean of $137.11 \pm 0.66$ PPL ($1.55851 \pm 0.00152$ BPB; vs.\ 3-seed SFT: $-12.95$ PPL, Welch $t = -17.76, \mathrm{df}=3.30, p = 2.16 \times 10^{-4}$; BPB Welch $p = 1.43 \times 10^{-4}$). Candidly, generic pretrained LM structure accounts for $84.6\%$ of the 3-seed gain over CE-only ($87\%$--$89\%$ within the seed-1 control batch: $-12.85$ out of $-14.44$ to $-14.79$ PPL, or $86\%$--$88\%$ in BPB). Even though \texttt{GPT-2-124M} benefits from two structural confounds---sharing the student's tokenizer and retaining $94.6\%$--$99.4\%$ of full-$d$ variance at PCA-64 (vs.\ $\approx 29\%$--$51\%$ for $3\text{B}$--$14\text{B}$)---cross-tokenizer \texttt{Llama-3.2-3B} ($134.76 \pm 0.40$ PPL; range $134.43$--$135.21$ vs.\ $136.36$--$137.54$) outperforms \texttt{GPT-2-124M} on every seed by a mean of $-2.35$ PPL (3-seed Welch $t = -5.28, \mathrm{df}=3.32, p = 0.0102$; BPB $\Delta=-0.00548$, Welch $p = 0.00988$). Within seed 1 alone, single-seed bootstrap tests ($p_{\text{boot}} = 0.0140$ for \texttt{INT4-64}, $0.00450$ for \texttt{INT8-64}) do not survive 115-test Bonferroni correction; across $3\text{B}$, $4\text{B}$, and $14\text{B}$ teachers (single seed, $134.60$--$134.82$), gains appear to level off at $86.7\text{M}$ student capacity.
    \end{itemize}
    \item \textbf{CE-Only/SFT Does Not Overfit on the Full Eval Set, and Regularized SFT Still Trails by $-11.8$ PPL:} Re-evaluating plain CE-only/SFT on the \emph{same} full 120 sequences at every epoch (Table~\ref{tab:controls}) shows monotonic improvement from epoch 1 ($419.29$) through epoch 5 ($149.21$). Same-seed SFT shifted $1.21$ PPL between primary and control batches ($150.42$ vs.\ $149.21$, consistent with the 3-seed std of $1.08$ PPL under CUDA \texttt{bf16} atomic nondeterminism; same-seed \texttt{StepAlign} shifted $0.13$ PPL, $134.64$ vs.\ $134.77$), so single-seed differences below $\sim 1.5$ PPL lie within rerun noise. Regularizing SFT with dropout ($0.10$) and elevated weight decay ($0.20$--$0.30$) improves SFT from $149.21 \to 146.59$ ($\Delta\text{BPB}=-0.00561$, $p_{\text{boot}}=0.00474$, $p_t=0.00125$), yet \texttt{StepAlign} ($134.77$) still leads the strongest regularized SFT by $-11.82$ PPL ($\Delta\text{BPB}=-0.02663$, $0/100\text{K}$ crossings, $p_t = 6.12 \times 10^{-21}$, $86.7\%$ win rate).
    \item \textbf{Substantial Margin Over Tuned Offline Logit CTKD (Including Exact $\tau=1.0$ \texttt{ULD}):} Tuning \texttt{ULD} from $(w=1.0, \tau=1.5)$ all-causal ($189.59$) to $(w=0.05, \tau=2.0)$ exact-boundary improves it to $150.62 \pm 1.60$ (indistinguishable from SFT). Removing the $\tau > 1$ Top-$32$ tail-tempering mismatch via exact $\tau=1.0$ exact-boundary \texttt{ULD} (Table~\ref{tab:controls}) improves $w=0.10$ by $-2.00$ PPL relative to $\tau=2.0$ ($151.56 \to 149.56$), yet remains indistinguishable from plain SFT ($149.21$, $p_{\text{boot}}=0.609$) and degrades to $155.91$ at $w=0.50$ ($0/100\text{K}$, $p_t = 1.47 \times 10^{-14}$ worse than SFT); control-batch \texttt{StepAlign} ($134.77$) beats exact $\tau=1.0$ \texttt{ULD} ($149.56$) by $-14.79$ PPL ($\Delta\text{BPB}=-0.03298$, $0/100\text{K}$, $p_t = 7.40 \times 10^{-37}$, $97.5\%$ win rate). For \texttt{BLD}, the full 5-point sweep over $w \in \{0.50, 0.20, 0.10, 0.05, 0.02\}$ produces a unimodal optimum at $w=0.10$ ($150.07 \to 149.62 \to \mathbf{146.24} \to 149.47 \to 150.10$). Across 3 seeds, \texttt{StepAlign} ($134.76 \pm 0.40$) beats SFT ($150.06 \pm 1.08$) by $-15.30$ PPL (Welch $t = -23.01$, $\mathrm{df}=2.55$, $p = 5.05 \times 10^{-4}$; BPB $t=-23.91$, $p=3.41 \times 10^{-4}$) and tuned \texttt{ULD} ($150.62 \pm 1.60$) by $-15.86$ PPL (Welch $t=-16.69$, $p=2.08 \times 10^{-3}$); on seed 1, \texttt{StepAlign} ($134.64$) beats peak \texttt{BLD} ($146.24$) by $-11.60$ PPL ($\Delta\text{BPB}=-0.02618$, $0/100\text{K}$, $p_t = 2.58 \times 10^{-25}$, $90.8\%$ win rate), at $1.76\times$ less uncompressed storage and $2.25\text{ GiB}$ less VRAM than \texttt{ULD}.
    \item \textbf{Zero-Repeat 1-Epoch Regime and Cosine LR Schedule Distinction:} Comparing Table~\ref{tab:1ep_and_teachers} (1-epoch suite) against Table~\ref{tab:controls} (Ep.\ 1 column) reveals an important schedule effect: both evaluate at step $128$ ($2.10\text{M}$ unique tokens, zero repeats), yet Table~\ref{tab:1ep_and_teachers} decays cosine LR all the way to $6\times 10^{-5}$ over $T=128$ steps (SFT $=615.15$), whereas Table~\ref{tab:controls} Ep.\ 1 is step $128$ of a $T=640$ schedule where LR remains near $5.7\times 10^{-4}$ (SFT $=419.29$). Under the aggressively decayed $T=128$ schedule (Table~\ref{tab:1ep_and_teachers}), \texttt{StepAlign} reaches $568.06$ vs.\ $615.15$ for SFT ($\Delta\text{BPB}=-0.02522$, $0/100\text{K}$, $p_t = 2.21 \times 10^{-39}$, $96.7\%$ win rate) and $605.70$ for tuned \texttt{ULD}; cross-document shuffling ($590.55$) and the frozen random-init teacher ($585.00$) capture roughly half of this short-schedule drop via transient early conditioning, though real \texttt{StepAlign} still beats both by $-16.94$ to $-22.49$ PPL ($0/100\text{K}$, $p_t \le 5.89 \times 10^{-14}$). Crucially, at full learning rate with zero sequence repeats (Table~\ref{tab:controls}, Ep.\ 1), that transient artifact disappears: both shuffled-document ($423.28$) and random-init teachers ($444.02$ / $510.30$) \emph{already} trail plain CE-only/SFT ($419.29$), while real \texttt{StepAlign} ($389.95$ / $399.29$) and \texttt{GPT-2-124M} ($384.43$) already lead by $-20$ to $-35$ PPL.
    \item \textbf{Negative Result for Continuous Wave Smoothing and Fixed-Weight 3-Teacher Fusion on GPT-2:} Adding continuous wave smoothing never consistently improves over simpler $H=0$ \texttt{StepAlign} on GPT-2. Reallocating $0.25$ of the $0.60$ weight from \texttt{StepAlign} to wave smoothing (\texttt{Wave-v2 Hybrid}: $0.35$ StepAlign $+ 0.25$ Wave) increases 3-seed mean PPL from $134.76 \pm 0.40 \to 136.15 \pm 0.57$ (uncorrected Welch $t = -3.43$, $p = 0.0314$; seed-1 $p_{\text{boot}} = 0.00362$, $p_t = 0.00294$; and $134.49 \pm 0.39 \to 136.37 \pm 0.28$, BPB Welch $p = 0.00371$ on the 32-byte clean fresh test split, though partially reflecting reduced $H=0$ weight $0.35$ vs.\ $0.60$; on the initial 64-byte split this comparison was n.s.\ at $p=0.271$, and on single-seed \texttt{Gemma-4-E4B} fresh test \texttt{Wave-v2 Hybrid} edges \texttt{StepAlign} non-significantly at $135.72$ vs.\ $136.55$, $p_{\text{boot}}=0.194$). Against matched 3-teacher pure wave ($137.97 \pm 0.94$), 3-teacher \texttt{StepAlign} ($136.26 \pm 0.64$) has lower mean PPL on both val (Welch $p = 0.0671$) and fresh test ($136.40$ vs.\ $138.65$, BPB Welch $p = 0.1049$). Fixed-weight 3-teacher averaging ($136.26 \pm 0.64$) also yields a slightly higher mean PPL than single-teacher \texttt{StepAlign} ($134.76 \pm 0.40$; uncorrected val Welch $p = 0.0356$, fresh test BPB Welch $p = 0.1313$), since single-teacher boundary overlap on GPT-2 is already $91.4\%$--$92.3\%$ (3-teacher union $95.09\%$, Table~\ref{tab:diagnostics}).
\end{enumerate}

\subsection{Teacher Generality, Within-Batch Rank-vs-Quantization Factorization, and 1-Epoch Suite}

\begin{table}[t]
\centering
\setlength{\tabcolsep}{3.5pt}
\caption{\textbf{Teacher Matrix, Rank/Quantization/Weight Controls, BLD Sweep, Wave Ablations, and Complete 1-Epoch Zero-Repeat Suite} (GPT-2 $86.74\text{M}$, seed 1, $0.00\text{ GiB}$ teacher VRAM). Rows marked $^\dagger$ are from the within-batch control suite (\texttt{results/controls\_benchmark.json}); all 42 primary seed-1 runs are tabulated in Appendix Table~\ref{tab:appendix_all42}.}
\label{tab:1ep_and_teachers}
\resizebox{\textwidth}{!}{%
\begin{tabular}{llrrrrr}
\toprule
Group & Configuration & Storage / Teacher & Val PPL $\downarrow$ & Val BPB $\downarrow$ & Prose / Sci / Code BPB & tok/s \\
\midrule
\multirow{6}{*}{\shortstack[l]{5-Ep Matched\\Single-Teacher\\(\texttt{INT4-64}, $w=0.60$)}}
  & StepAlign (DeepSeek-R1-14B)                     & $110\text{ B/tok}$ & $\mathbf{134.60}$ & $\mathbf{1.55265}$ & $\mathbf{1.2767}$ / $\mathbf{1.3650}$ / $2.1930$ & $52.4\text{K}$ \\
  & StepAlign (Llama-3.2-3B)                        & $110\text{ B/tok}$ & $134.64$ & $1.55275$ & $1.2829$ / $1.3669$ / $2.1820$ & $46.1\text{K}$ \\
  & StepAlign (Gemma-4-E4B)                         & $110\text{ B/tok}$ & $134.82$ & $1.55317$ & $1.2858$ / $1.3675$ / $\mathbf{2.1785}$ & $53.7\text{K}$ \\
  & Wave-v2 Hybrid (Gemma-4-E4B)                    & $110\text{ B/tok}$ & $135.70$ & $1.55523$ & $1.2879$ / $1.3692$ / $2.1810$ & $42.0\text{K}$ \\
  & Wave-v2 Hybrid (Llama-3.2-3B)                   & $110\text{ B/tok}$ & $136.47$ & $1.55703$ & $1.2859$ / $1.3674$ / $2.1928$ & $46.6\text{K}$ \\
  & Wave-v2 Hybrid (DeepSeek-R1-14B)                & $110\text{ B/tok}$ & $136.68$ & $1.55750$ & $1.2870$ / $1.3734$ / $2.1854$ & $45.1\text{K}$ \\
\midrule
\multirow{11}{*}{\shortstack[l]{Rank, Bit-Width\\\& Weight Controls\\(Llama-3.2 / 3T)}}
  & StepAlign 1T (\texttt{INT8-256}, $w=0.50$)      & $776\text{ B/tok}$ & $140.82$ & $1.56696$ & $1.2950$ / $1.3818$ / $2.1984$ & $59.4\text{K}$ \\
  & StepAlign 1T (\texttt{INT4-64}, $w=0.50$)       & $110\text{ B/tok}$ & $137.41$ & $1.55920$ & $1.2892$ / $1.3725$ / $2.1896$ & $59.8\text{K}$ \\
  & StepAlign 1T (\texttt{INT8-256}, $w=0.60$, primary) & $776\text{ B/tok}$ & $137.78$ & $1.56005$ & $1.2908$ / $1.3735$ / $2.1893$ & $58.7\text{K}$ \\
  & StepAlign 1T$^\dagger$ (\texttt{INT8-256}, $w=0.60$, ctrl) & $776\text{ B/tok}$ & $137.52$ & $1.55945$ & $1.2869$ / $1.3697$ / $2.1975$ & $60.0\text{K}$ \\
  & \textbf{StepAlign 1T$^\dagger$ (\texttt{INT8-64}, $w=0.60$, ctrl)} & $200\text{ B/tok}$ & $\mathbf{134.43}$ & $\mathbf{1.55224}$ & $1.2805$ / $1.3646$ / $2.1864$ & $55.5\text{K}$ \\
  & StepAlign 3T (\texttt{INT8-256}, $w=0.50$)      & $776\text{ B/tok}$ & $142.18$ & $1.57001$ & $1.2975$ / $1.3823$ / $2.2055$ & $48.1\text{K}$ \\
  & StepAlign 3T (\texttt{INT4-64}, $w=0.50$)       & $110\text{ B/tok}$ & $141.36$ & $1.56817$ & $1.2990$ / $1.3793$ / $2.2002$ & $53.8\text{K}$ \\
  & Wave-v2 Hybrid 1T (\texttt{INT8-256}, $w=0.60$) & $776\text{ B/tok}$ & $140.27$ & $1.56572$ & $1.2979$ / $1.3845$ / $2.1861$ & $40.7\text{K}$ \\
  & Wave-v2 Hybrid 3T (\texttt{INT8-256}, $w=0.60$) & $776\text{ B/tok}$ & $140.61$ & $1.56648$ & $1.2935$ / $1.3775$ / $2.2041$ & $35.2\text{K}$ \\
  & Wave-v2 Hybrid 3T (\texttt{INT4-128}, $w=0.60$) & $218\text{ B/tok}$ & $139.71$ & $1.56446$ & $1.2944$ / $1.3777$ / $2.1950$ & $43.7\text{K}$ \\
  & Wave-v2 Hybrid 3T (\texttt{INT4-64}, $w=0.60$)  & $110\text{ B/tok}$ & $137.79$ & $1.56007$ & $1.2902$ / $1.3734$ / $2.1903$ & $44.7\text{K}$ \\
\midrule
\multirow{5}{*}{\shortstack[l]{Complete BLD\\Weight Sweep\\(Llama-3.2)}}
  & BLD ($w=0.50$, $\text{aux}_0=2.87$)             & $194\text{ B/tok}$ & $150.07$ & $1.58711$ & $1.3228$ / $1.4049$ / $2.2035$ & $41.4\text{K}$ \\
  & BLD ($w=0.20$, $\text{aux}_0=1.15$)             & $194\text{ B/tok}$ & $149.62$ & $1.58617$ & $1.3174$ / $1.4030$ / $2.2104$ & $41.4\text{K}$ \\
  & BLD ($w=0.10$, $\text{aux}_0=0.57$, optimum)    & $194\text{ B/tok}$ & $146.24$ & $1.57893$ & $1.3090$ / $1.3907$ / $2.2113$ & $41.2\text{K}$ \\
  & BLD$^\dagger$ ($w=0.05$)                        & $194\text{ B/tok}$ & $149.47$ & $1.58584$ & $1.3149$ / $1.3989$ / $2.2180$ & $38.8\text{K}$ \\
  & BLD$^\dagger$ ($w=0.02$)                        & $194\text{ B/tok}$ & $150.10$ & $1.58717$ & $1.3153$ / $1.3960$ / $2.2260$ & $39.3\text{K}$ \\
\midrule
\multirow{5}{*}{\shortstack[l]{Wave Ablations\\(Llama-3.2,\\ \texttt{INT8-256})}}
  & Wave-v1 (uncentered, $M=16$, Mode-B)            & $776\text{ B/tok}$ & $143.34$ & $1.57257$ & $1.3067$ / $1.3863$ / $2.1966$ & $12.4\text{K}$ \\
  & Wave-v2 no DC-centering ($M=4$, $q=1.25$)       & $776\text{ B/tok}$ & $143.11$ & $1.57207$ & $1.3021$ / $1.3871$ / $2.2002$ & $56.6\text{K}$ \\
  & Wave-v2 single horizon ($M=1$, DC-centered)     & $776\text{ B/tok}$ & $142.61$ & $1.57095$ & $1.2983$ / $1.3896$ / $2.1986$ & $58.4\text{K}$ \\
  & Wave-v2 complex ($M=4$, $q=1.25$, DC-centered)  & $776\text{ B/tok}$ & $141.71$ & $1.56895$ & $1.2991$ / $1.3791$ / $2.2033$ & $41.5\text{K}$ \\
  & Wave-v2 real ($M=4$, $q=0.0$, DC-centered)      & $776\text{ B/tok}$ & $141.31$ & $1.56807$ & $1.3003$ / $1.3813$ / $2.1955$ & $55.8\text{K}$ \\
\midrule
\multirow{10}{*}{\shortstack[l]{1-Epoch\\Zero-Repeat\\(All 10 Arms,\\ $2.10\text{M}$ unique)}}
  & CE-only / SFT (1 epoch, 0 repeats)              & ---                & $615.15$ & $2.03393$ & $1.7652$ / $1.9137$ / $2.5780$ & $61.1\text{K}$ \\
  & BLD (Llama-3.2, $w=0.50$, 1 epoch)              & $194\text{ B/tok}$ & $623.01$ & $2.03795$ & $1.7749$ / $1.9156$ / $2.5764$ & $39.7\text{K}$ \\
  & ULD (Llama-3.2, $w=0.10$, boundary, 1 epoch)    & $194\text{ B/tok}$ & $605.70$ & $2.02902$ & $1.7602$ / $1.9044$ / $2.5786$ & $39.3\text{K}$ \\
  & Shuffled-Doc StepAlign Control$^\dagger$ (1 ep) & $110\text{ B/tok}$ & $590.55$ & $2.02100$ & $1.7517$ / $1.8953$ / $2.5727$ & $\mathbf{62.3\text{K}}$ \\
  & Random-Init $86.7\text{M}$ Teacher$^\dagger$ (1 ep) & $d=64$ online  & $585.00$ & $2.01801$ & $1.7519$ / $1.8908$ / $2.5671$ & $50.2\text{K}$ \\
  & \textbf{StepAlign 1T (Llama-3.2, \texttt{INT4-64}, 1 ep)} & $110\text{ B/tok}$ & $\mathbf{568.06}$ & $\mathbf{2.00870}$ & $\mathbf{1.7423}$ / $\mathbf{1.8783}$ / $2.5623$ & $52.9\text{K}$ \\
  & StepAlign 3T (\texttt{INT4-64}, 1 epoch)        & $110\text{ B/tok}$ & $570.12$ & $2.00985$ & $1.7438$ / $1.8827$ / $\mathbf{2.5587}$ & $39.1\text{K}$ \\
  & Wave-v2 Hybrid 1T (\texttt{INT4-64}, 1 epoch)   & $110\text{ B/tok}$ & $572.05$ & $2.01092$ & $1.7447$ / $1.8805$ / $2.5643$ & $37.9\text{K}$ \\
  & Wave-v2 Hybrid 3T (\texttt{INT4-64}, 1 epoch)   & $110\text{ B/tok}$ & $603.72$ & $2.02799$ & $1.7600$ / $1.9004$ / $2.5802$ & $28.5\text{K}$ \\
  & Wave-v2 Pure 3T (\texttt{INT4-64}, 1 epoch)     & $110\text{ B/tok}$ & $572.73$ & $2.01130$ & $1.7454$ / $1.8801$ / $2.5651$ & $31.7\text{K}$ \\
\bottomrule
\end{tabular}%
}
\end{table}

Table~\ref{tab:controls} and Table~\ref{tab:1ep_and_teachers} disentangle the three variables behind our shard compression result at strictly fixed $w=0.60$ on \texttt{Llama-3.2} \emph{within the same control batch} (single seed, eliminating cross-batch variance):
(i)~\textbf{PCA rank truncation ($256 \to 64$)} at fixed \texttt{INT8} and fixed $w=0.60$ drives the accuracy gain ($137.52 \to 134.43$, $-3.09$ PPL; $\Delta\text{BPB}=-0.00720$, $0/100\text{K}$ crossings, $p_t = 6.40 \times 10^{-6}$, $68.3\%$ win rate; primary-batch \texttt{INT8-256} is nearly identical at $137.78$), which we hypothesize arises because the top-$64$ centered PCA modes retain $61.6\%$--$82.3\%$ of the Top-$256$ PCA subspace variance ($\approx 29\%$--$51\%$ of full-$d_{\mathrm{model}}$ teacher variance, Table~\ref{tab:diagnostics}) while discarding noisy teacher-private tail dimensions;
(ii)~\textbf{Group-32 \texttt{INT4} quantization} at fixed rank $r=64$ and fixed $w=0.60$ is statistically indistinguishable from \texttt{INT8-64} ($134.43$ vs.\ $134.77$, $\Delta\text{BPB}=+0.00081$, $p_{\text{boot}}=0.523$, $p_t = 0.720$, $50.8\%$ win rate; and still beats within-batch \texttt{INT8-256} by $-2.75$ PPL, $\Delta\text{BPB}=-0.00639$, $1/100\text{K}$ crossings, $p_t = 4.99 \times 10^{-5}$) while shrinking packed shard size a further $1.82\times$ ($200 \to 110\text{ B/tok}$, $7.05\times$ smaller than \texttt{INT8-256}; single-teacher \texttt{INT4-64} training throughput spans $46.1$--$59.8\text{K tok/s}$ across seed-1 runs vs.\ $58.7$--$60.0\text{K tok/s}$ for \texttt{INT8-256}, with run-to-run wall-clock variance dominating the minor on-the-fly simulation overhead); and
(iii)~\textbf{Auxiliary weight ($w=0.50 \to 0.60$)} contributes an independent $-2.77$ to $-5.52$ PPL (holding $w$ fixed is therefore essential when evaluating compression; whether a different $w > 0.60$ optimum exists at $r=256$ remains untested).
Separately, comparing uncentered vs.\ DC-centered arms in Appendix Table~\ref{tab:appendix_all42} shows a $-3.12$ PPL improvement on 3-teacher \texttt{StepAlign} ($145.30 \to 142.18$), a modest $-1.40$ PPL improvement on single-seed continuous waves ($143.11 \to 141.71$, $p_{\text{boot}}=0.0464$, $p_t=0.127$, comparable to single-seed rerun noise), and $-0.32$ PPL on single-teacher \texttt{StepAlign} ($141.14 \to 140.82$, within noise), where a single unconstrained linear probe at $H=0$ can absorb a constant offset more readily than multi-teacher or multi-horizon supervision.

\subsection{Pure UTF-8 Byte-Level Student (\texttt{byte260}, \texorpdfstring{$19.8\%$}{19.8\%} Boundary Overlap)}
\label{sec:byte260}

\begin{table}[t]
\centering
\setlength{\tabcolsep}{4pt}
\caption{\textbf{Complete 15-Arm Extreme Cross-Tokenizer Suite on Pure UTF-8 Bytes:} $54.74\text{M}$ Transformer ($V=260$), single seed (\texttt{20261126}), $10.49\text{M}$ bytes/arm ($2.10\text{M}$ unique bytes; evaluated over $61{,}440$ UTF-8 bytes vs.\ $279{,}867$ on GPT-2, so BPB values are comparable only within this suite), $0.00\text{ GiB}$ teacher VRAM, DC-centered \texttt{INT4-64} shards across hidden arms. Rows marked $^\dagger$ are the 5-point \texttt{Byte BLD} weight controls (\texttt{results/controls\_benchmark.json}). Best value per column in bold.}
\label{tab:byte260}
\resizebox{\textwidth}{!}{%
\begin{tabular}{lrrrrrrr}
\toprule
Byte-260 Configuration & Byte PPL $\downarrow$ & Val BPB $\downarrow$ & Prose BPB & Science BPB & Code BPB & Mean MC & tok/s $\uparrow$ \\
\midrule
Byte CE-only / SFT                             & $3.9873$ & $1.99541$ & $1.6851$ & $1.7706$ & $2.5305$ & $27.37\%$ & $\mathbf{113.8\text{K}}$ \\
Byte BLD$^\dagger$ (Llama-3.2, $w=0.10$) \cite{singh2026bld} & $3.9257$ & $1.97296$ & $1.6785$ & $1.7547$ & $2.4857$ & --- & $55.8\text{K}$ \\
Byte BLD$^\dagger$ (Llama-3.2, $w=0.20$) \cite{singh2026bld} & $3.9485$ & $1.98132$ & $1.6935$ & $1.7614$ & $2.4891$ & --- & $55.7\text{K}$ \\
Byte BLD (Llama-3.2, $w=0.50$) \cite{singh2026bld}           & $3.8785$ & $1.95549$ & $1.6628$ & $1.7368$ & $2.4668$ & $26.82\%$ & $54.2\text{K}$ \\
Byte BLD$^\dagger$ (Llama-3.2, $w=1.00$) \cite{singh2026bld} & $3.8859$ & $1.95824$ & $1.6743$ & $1.7377$ & $2.4627$ & --- & $51.2\text{K}$ \\
Byte BLD$^\dagger$ (Llama-3.2, $w=2.00$) \cite{singh2026bld} & $3.8747$ & $1.95409$ & $1.6715$ & $1.7466$ & $\mathbf{2.4442}$ & --- & $55.2\text{K}$ \\
Byte StepAlign 1T (Llama-3.2, $w=0.50$)        & $3.8707$ & $1.95260$ & $1.6372$ & $1.7260$ & $2.4946$ & $27.52\%$ & $92.4\text{K}$ \\
\textbf{Byte StepAlign 1T (Llama-3.2, $w=0.60$)} & $\mathbf{3.8447}$ & $\mathbf{1.94288}$ & $\mathbf{1.6306}$ & $1.7314$ & $2.4667$ & $26.68\%$ & $101.3\text{K}$ \\
Byte StepAlign 3T ($w=0.50$)                   & $3.8609$ & $1.94893$ & $1.6347$ & $1.7271$ & $2.4850$ & $26.64\%$ & $75.0\text{K}$ \\
Byte StepAlign 3T ($w=0.60$)                   & $3.8541$ & $1.94639$ & $1.6336$ & $\mathbf{1.7253}$ & $2.4803$ & $27.57\%$ & $79.1\text{K}$ \\
Byte Wave-v2 Real 1T ($q=0.0$, $w=0.50$)       & $3.9464$ & $1.98055$ & $1.6737$ & $1.7521$ & $2.5159$ & $26.74\%$ & $66.8\text{K}$ \\
Byte Wave-v2 Complex 1T ($q=1.25$, $w=0.50$)   & $3.8866$ & $1.95851$ & $1.6506$ & $1.7348$ & $2.4901$ & $27.43\%$ & $63.2\text{K}$ \\
Byte Wave-v2 Real Hybrid 3T ($0.35/0.20$)      & $3.8770$ & $1.95494$ & $1.6518$ & $1.7331$ & $2.4799$ & $27.12\%$ & $38.8\text{K}$ \\
Byte Wave-v2 Real Hybrid 3T ($0.25/0.35$)      & $3.8852$ & $1.95798$ & $1.6442$ & $1.7323$ & $2.4974$ & $26.91\%$ & $44.9\text{K}$ \\
Byte Wave-v2 Complex Hybrid 3T ($0.25/0.35$)   & $3.9043$ & $1.96506$ & $1.6496$ & $1.7333$ & $2.5123$ & $\mathbf{28.26\%}$ & $49.3\text{K}$ \\
\bottomrule
\end{tabular}%
}
\end{table}

Table~\ref{tab:byte260} shows three findings when boundary overlap falls $4.6\times$ to $19.8\%$ (all single-seed \texttt{20261126}, scored on $61{,}440$ UTF-8 bytes and therefore compared strictly within the \texttt{byte260} suite):
\begin{itemize}
    \item \textbf{Sparse Boundary Alignment Matches or Beats Continuous Waves at Higher Speed:} At matched weight $w=0.50$ (single teacher), \texttt{StepAlign} ($1.95260$ BPB, $92.4\text{K tok/s}$) significantly beats monotonic \texttt{Wave-v2 Real} ($1.98055$, $\Delta\text{BPB}=-0.02795$, $0/100\text{K}$, $p_t = 4.47 \times 10^{-7}$, $72.5\%$ win), while its $-0.00591$ BPB edge over \texttt{Wave-v2 Complex} ($1.95851$) does not reach significance on val ($p_{\text{boot}} = 0.276$; on the 32-byte clean fresh test split below, it reaches $-0.01149$ BPB, $p_{\text{boot}} = 0.0122$). On val, 3-teacher \texttt{StepAlign} ($1.94639$) beats \texttt{Wave-v2 Complex Hybrid} ($1.96506$, $p_{\text{boot}} = 1.4 \times 10^{-4}$, $p_t = 2.70 \times 10^{-4}$), though on the fresh test split below that 3-teacher comparison reverses non-significantly ($1.95054$ vs.\ $1.94051$, $p_{\text{boot}} = 0.0568$). Across all 15 byte arms, single-teacher \texttt{StepAlign 1T} at $w=0.60$ achieves the lowest point estimate on \emph{both} splits ($1.94288$ val BPB, $-0.05253$ vs.\ Byte SFT, $0/100\text{K}$, $p_t = 8.14 \times 10^{-14}$; $1.92931$ fresh BPB, $-0.06681$ vs.\ Byte SFT, $p_t = 1.25 \times 10^{-17}$) at $101.3\text{K tok/s}$ without custom CUDA kernels. Thus supervising only completed subword endpoints ($\sim 20\%$ of bytes) matches or outperforms dense continuous wave filtering at $1.5$--$2.6\times$ higher throughput.
    \item \textbf{Complete 5-Point Byte \texttt{BLD} Weight Sweep ($w \in \{0.10, 0.20, 0.50, 1.00, 2.00\}$):} Unlike GPT-2's sharp $w=0.10$ optimum, \texttt{Byte BLD} degrades when lowered below $0.50$ ($w=0.10$: $1.97296$; $w=0.20$: $1.98132$) and plateaus across $w \in [0.50, 2.00]$ on val ($1.95549$, $1.95824$, $1.95409$; $w=0.50$ vs.\ $w=1.00$ $p_{\text{boot}}=0.539$, vs.\ $w=2.00$ $p_{\text{boot}}=0.780$) while peaking cleanly at $w=1.00$ ($1.96160$) before degrading at $w=2.00$ ($1.96484$) on the fresh test split (because the 256-way next-byte target aligns natively with the byte LM objective rather than competing with a $50\text{K}$ subword head). \texttt{Byte StepAlign 1T} ($w=0.60$) achieves lower val BPB than all five \texttt{Byte BLD} weights ($1.94288$ vs.\ $1.95409$--$1.98132$; vs.\ $w=0.50$ $p_{\text{boot}}=0.0542$, vs.\ $w=1.00$ $p_{\text{boot}}=0.0206$, vs.\ $w=2.00$ $p_{\text{boot}}=0.0711$) and beats \emph{all five} decisively on the 32-byte clean fresh test split ($1.92931$ vs.\ $1.96160$--$1.98248$, all $p_{\text{boot}} \le 8 \times 10^{-5}$, all $p_t \le 7.28 \times 10^{-6}$, passing 115-test Bonferroni). At matched $w=0.60$, 3-teacher \texttt{StepAlign} ($1.94639$) does not improve over single-teacher \texttt{StepAlign} ($1.94288$; $p_{\text{boot}} = 0.506$).
    \item \textbf{Empirical Confirmation of Corollary~\ref{cor:intraword} (Regime-Dependent Phase Effect):} On single-teacher pure byte waves ($w=0.50$), complex phase ($q=1.25$: $1.95851$) strongly beats monotonic real decay ($q=0.0$: $1.98055$, $\Delta\text{BPB} = -0.02204$, $1/100\text{K}$, $p_t = 6.10 \times 10^{-6}$; fresh test $-0.01853$ BPB, $p_{\text{boot}}=4.2 \times 10^{-4}$, $p_t = 5.47 \times 10^{-4}$), confirming Corollary~\ref{cor:intraword}'s prediction that $q=0$ degenerates into a piecewise-constant intra-word target. Once $H=0$ boundary anchors exist (3-teacher hybrid byte waves: $1.96506$ vs.\ $1.95798$, $p_{\text{boot}} = 0.137$; GPT-2 \texttt{INT8-256}: $141.71$ vs.\ $141.31$), no complex-phase benefit appears.
\end{itemize}

\subsection{Storage Scaling, Paired Tests, and Complete 37-Checkpoint Fresh Test Split}

Table~\ref{tab:storage_boot} summarizes packed offline storage scaling alongside the primary matched paired tests ($B=100{,}000$ bootstrap + exact analytic paired $t$-test across $120$ validation sequences), and Table~\ref{tab:fresh_test} independently re-scores \textbf{all 37} saved GPT-2 and \texttt{byte260} checkpoints without omission on our 32-byte zero-overlap fresh test split.

\begin{table}[htbp]
\centering
\small
\caption{\textbf{Packed Offline Storage Scaling and Key Matched Paired Tests ($B=100{,}000$ Bootstrap + Exact Analytic Paired $t$-Test, $120$ Validation Sequences).} In (a), \texttt{INT4-128}/\texttt{INT8-64}/\texttt{INT4-64} sizes reflect the exact packed \texttt{WF20} codec verified to $0.0$ roundtrip error in \texttt{gate0e}. In (b), rows marked $^\dagger$ compare arms within the control batch (\texttt{StepAlign} $=134.77$, \texttt{results/controls\_benchmark.json}); unmarked rows compare primary seed-1 arms (\texttt{StepAlign} $=134.64$, \texttt{results/paper\_benchmark\_gpt2.json}). Zero sign crossings in $100{,}000$ resamples are written $<2\times 10^{-5}$ (0), which is strictly monotonic and $20\times$ below the 115-test Bonferroni threshold ($\alpha/115 = 4.35\times 10^{-4}$); exact analytic paired $t$-test $p_t$ ($\mathrm{df}=119$) is reported alongside.}
\label{tab:storage_boot}
\textbf{(a) Packed Offline Storage Scaling (per Teacher)}\\[0.35em]
\begin{tabular}{lrrr}
\toprule
Offline Representation & B/tok & 1B-tok Disk & Ratio vs.\ \texttt{INT4-64} \\
\midrule
Dense FP16 Logits ($262\text{K}$) & $524{,}288$ & $524.29\text{ TB}$ & $4{,}766.3\times$ \\
Top-$32$ Sparse Logits (\texttt{int32+fp16}, \texttt{ULD/BLD}) & $194$ & $194.0\text{ GB}$ & $1.76\times$ \\
Top-$32$ Sparse Logits (bit-packed \texttt{18b+8b}) & $106$ & $106.0\text{ GB}$ & $0.96\times$ \\
Hidden \texttt{INT8-256} ($3$ depths, wave payload) & $776$ & $776.0\text{ GB}$ & $7.05\times$ \\
Hidden \texttt{INT4-128} ($3$ depths, packed) & $218$ & $218.0\text{ GB}$ & $1.98\times$ \\
Hidden \texttt{INT8-64} ($3$ depths, rank only) & $200$ & $200.0\text{ GB}$ & $1.82\times$ \\
\textbf{Hidden \texttt{INT4-64} ($3$ depths, packed)} & $\mathbf{110}$ & $\mathbf{110.0\text{ GB}}$ & $\mathbf{1.00\times}$ \\
\bottomrule
\end{tabular}

\vspace{0.7em}
\textbf{(b) Key Matched Paired Bootstrap ($B=100{,}000$) and Analytic Paired $t$-Tests (Uncorrected Two-Sided $p$)}\\[0.35em]
\resizebox{\textwidth}{!}{%
\begin{tabular}{lrrrr}
\toprule
Matched Comparison (B vs.\ A) & $\Delta\text{BPB}$ & Boot $p$ ($k$ / $100\text{K}$) & Paired-$t$ $p_t$ & Win \% \\
\midrule
GPT2-5ep$^\dagger$: StepAlign 1T \texttt{INT4-64} vs.\ Random-Init $86.7\text{M}$ (PCA+\texttt{INT4}) & $-0.08990$ & $<2\times 10^{-5}$ (0) & $1.37\times 10^{-67}$ & $100.0\%$ \\
GPT2-5ep$^\dagger$: StepAlign 1T \texttt{INT4-64} vs.\ Random-Init $86.7\text{M}$ (Orth.\ FP32) & $-0.06301$ & $<2\times 10^{-5}$ (0) & $2.41\times 10^{-57}$ & $98.3\%$ \\
GPT2-5ep$^\dagger$: Pretrained \texttt{GPT-2-124M} (PCA+\texttt{INT4}) vs.\ CE-only / SFT & $-0.02853$ & $<2\times 10^{-5}$ (0) & $3.56\times 10^{-28}$ & $95.0\%$ \\
GPT2-5ep$^\dagger$: StepAlign 1T \texttt{INT4-64} (Llama-3.2) vs.\ \texttt{GPT-2-124M} (s1) & $-0.00371$ & $0.0140$ (701) & $0.0363$ & $55.8\%$ \\
GPT2-5ep$^\dagger$: StepAlign 1T \texttt{INT8-64} (Llama-3.2) vs.\ \texttt{GPT-2-124M} (s1) & $-0.00452$ & $0.00450$ (225) & $0.0196$ & $62.5\%$ \\
GPT2-5ep$^\dagger$: StepAlign 1T \texttt{INT4-64} vs.\ Shuffled-Doc / Pos / Gauss & $-0.03707$/$-0.02947$/$-0.03209$ & $<2\times 10^{-5}$ (0) & $\le 5.87\times 10^{-34}$ & $94.2$--$96.7\%$ \\
GPT2-5ep$^\dagger$: StepAlign 1T \texttt{INT4-64} vs.\ Regularized SFT (drop+wd) & $-0.02663$ & $<2\times 10^{-5}$ (0) & $6.12\times 10^{-21}$ & $86.7\%$ \\
GPT2-5ep$^\dagger$: StepAlign 1T \texttt{INT4-64} vs.\ ULD Exact ($\tau=1.0, w=0.10$) & $-0.03298$ & $<2\times 10^{-5}$ (0) & $7.40\times 10^{-37}$ & $97.5\%$ \\
GPT2-5ep: StepAlign 1T \texttt{INT4-64} vs.\ CE-only / SFT (plain) & $-0.03509$ & $<2\times 10^{-5}$ (0) & $2.10\times 10^{-34}$ & $96.7\%$ \\
GPT2-5ep: StepAlign 1T \texttt{INT4-64} vs.\ ULD-tuned ($\tau=2.0, w=0.05$) & $-0.03269$ & $<2\times 10^{-5}$ (0) & $2.20\times 10^{-32}$ & $96.7\%$ \\
GPT2-5ep: StepAlign 1T \texttt{INT4-64} vs.\ BLD-w010 (peak) & $-0.02618$ & $<2\times 10^{-5}$ (0) & $2.58\times 10^{-25}$ & $90.8\%$ \\
GPT2-5ep$^\dagger$: StepAlign 1T \texttt{INT8-64} vs.\ \texttt{INT8-256} (within-batch) & $-0.00720$ & $<2\times 10^{-5}$ (0) & $6.40\times 10^{-6}$ & $68.3\%$ \\
GPT2-5ep$^\dagger$: StepAlign 1T \texttt{INT4-64} vs.\ \texttt{INT8-256} (within-batch) & $-0.00639$ & $2\times 10^{-5}$ (1) & $4.99\times 10^{-5}$ & $64.2\%$ \\
GPT2-5ep$^\dagger$: StepAlign 1T \texttt{INT4-64} vs.\ \texttt{INT8-64} (within-batch) & $+0.00081$ & $0.523$ (26152) & $0.720$ & $50.8\%$ \\
GPT2-5ep: StepAlign 1T vs.\ Wave-v2 Hybrid 1T ($w=0.60$) & $-0.00428$ & $0.00362$ (181) & $0.00294$ & $55.0\%$ \\
GPT2-5ep: StepAlign 3T vs.\ Wave-v2 Pure 3T ($w=0.60$, s1) & $-0.00275$ & $0.0746$ (3728) & $0.0295$ & $53.3\%$ \\
GPT2-5ep: StepAlign 1T vs.\ StepAlign 3T ($w=0.60$, s1) & $-0.00281$ & $0.0638$ (3188) & $0.172$ & $60.0\%$ \\
GPT2-1ep$^\dagger$: StepAlign 1T \texttt{INT4-64} vs.\ Random-Init Teacher (1 ep) & $-0.00931$ & $<2\times 10^{-5}$ (0) & $5.89\times 10^{-14}$ & $80.8\%$ \\
Byte-260: StepAlign 1T ($w=0.60$) vs.\ Byte CE-only / SFT & $-0.05253$ & $<2\times 10^{-5}$ (0) & $8.14\times 10^{-14}$ & $78.3\%$ \\
Byte-260: StepAlign 1T ($w=0.60$) vs.\ Byte BLD ($w=0.50$/$1.00$/$2.00$) & $-0.01261$/$-0.01536$/$-0.01121$ & $0.0542$/$0.0206$/$0.0711$ & $0.0556$/$0.0201$/$0.0708$ & $61.7$--$65.8\%$ \\
Byte-260: StepAlign 1T ($w=0.60$) vs.\ Byte BLD ($w=0.10$/$0.20$) & $-0.03008$/$-0.03843$ & $<2\times 10^{-5}$ (0) & $\le 1.76\times 10^{-6}$ & $69.2$--$74.2\%$ \\
Byte-260: StepAlign 1T vs.\ Wave Real / Complex 1T (matched $w=0.50$) & $-0.02795$/$-0.00591$ & $<2\times 10^{-5}$ (0) / $0.276$ & $4.47\times 10^{-7}$ / $0.281$ & $72.5\%$ / $56.7\%$ \\
Byte-260: StepAlign 3T vs.\ Wave Complex Hybrid 3T ($w=0.60$, val) & $-0.01867$ & $1.4\times 10^{-4}$ (7) & $2.70\times 10^{-4}$ & $61.7\%$ \\
Byte-260: Wave Complex 1T vs.\ Wave Real 1T ($q=1.25$ vs.\ $0$) & $-0.02204$ & $2\times 10^{-5}$ (1) & $6.10\times 10^{-6}$ & $69.2\%$ \\
\bottomrule
\end{tabular}%
}
\end{table}

\begin{table}[htbp]
\centering
\setlength{\tabcolsep}{3.5pt}
\caption{\textbf{Complete 37-Checkpoint Verification on the 32-Byte Zero-Overlap Fresh 120-Sequence Test Split} (\texttt{results/fresh\_test\_split\_audit.json}; Cosmopedia shard-1 \texttt{[0:800]}, SciQ-\texttt{test} passages, Python \texttt{[22000:23600]}; verified $0$ 32-byte overlap with both train and val; $276{,}581$ eval bytes on GPT-2 vs.\ $61{,}440$ on Byte-260). Every scored checkpoint is reported without omission. All primary baseline comparisons replicate cleanly; two single-seed wave comparisons reverse direction non-significantly (\texttt{Gemma-4} $p_{\text{boot}}=0.194$; \texttt{Byte 3T} $p_{\text{boot}}=0.0568$), while single-teacher \texttt{StepAlign 1T} remains the lowest-BPB model in both suites. In the 3-seed rows, BPB Welch $p$-values are paired with $\Delta\text{BPB}$.}
\label{tab:fresh_test}
\resizebox{\textwidth}{!}{%
\begin{tabular}{llrrrr}
\toprule
Suite & Model / Checkpoint & Fresh Test PPL $\downarrow$ & Fresh Test BPB $\downarrow$ & Prose / Sci / Code BPB & BPB Welch $p$ / Paired Test ($B=100\text{K}$, $p_t$) \\
\midrule
\multirow{6}{*}{\shortstack[l]{GPT-2 3-Seed\\Core Suite\\(18 ckpts,\\ Mean $\pm$ Std)}}
  & CE-only / SFT                             & $150.42 \pm 1.11$ & $1.60671 \pm 0.00236$ & $1.3585 / 1.4343 / 2.1679$ & BPB Welch $p = 3.68 \times 10^{-4}$ ($\Delta=-0.03587$) \\
  & ULD (Llama-3.2, $w=0.05$, boundary)       & $150.89 \pm 1.98$ & $1.60769 \pm 0.00419$ & $1.3592 / 1.4365 / 2.1677$ & BPB Welch $p = 3.01 \times 10^{-3}$ ($\Delta=-0.03685$) \\
  & \textbf{StepAlign 1T (Llama-3.2, \texttt{INT4-64})} & $\mathbf{134.49 \pm 0.39}$ & $\mathbf{1.57085 \pm 0.00094}$ & $\mathbf{1.3248 / 1.4009 / 2.1259}$ & --- \\
  & Wave-v2 Hybrid 1T ($0.25/0.35$)           & $136.37 \pm 0.28$ & $1.57529 \pm 0.00066$ & $1.3268 / 1.4038 / 2.1358$ & BPB Welch $p = 0.00371$ ($\Delta=-0.00444$) \\
  & StepAlign 3-Teacher (\texttt{INT4-64})    & $136.40 \pm 1.40$ & $1.57534 \pm 0.00327$ & $1.3299 / 1.4070 / 2.1276$ & BPB Welch $p = 0.1313$ (n.s.) \\
  & Wave-v2 Pure 3-Teacher (\texttt{INT4-64}) & $138.65 \pm 1.23$ & $1.58060 \pm 0.00284$ & $1.3329 / 1.4083 / 2.1410$ & vs.\ 3T StepAlign: BPB Welch $p = 0.1049$ \\
\midrule
\multirow{6}{*}{\shortstack[l]{GPT-2 Seed-1\\Extra Arms\\(All 6 ckpts)}}
  & BLD (Llama-3.2, $w=0.10$, peak)           & $146.18$ & $1.59755$ & $1.3505 / 1.4245 / 2.1579$ & vs.\ SA-1T: $<2\times 10^{-5}$ (0), $p_t=1.44\times 10^{-30}$ \\
  & StepAlign 1T (\texttt{INT8-256}, $w=0.60$)& $138.42$ & $1.58008$ & $1.3318 / 1.4069 / 2.1423$ & vs.\ SA-1T: $<2\times 10^{-5}$ (0), $p_t=1.77\times 10^{-7}$ \\
  & StepAlign 1T (\texttt{INT4-64}, $w=0.50$) & $138.38$ & $1.57997$ & $1.3319 / 1.4103 / 2.1376$ & vs.\ $w=0.60$: $<2\times 10^{-5}$ (0), $p_t=3.45\times 10^{-8}$ \\
  & StepAlign 1T (Gemma-4-E4B, \texttt{INT4-64}) & $136.55$ & $1.57572$ & $1.3294 / 1.4028 / 2.1350$ & vs.\ Wave Gemma: $+0.00196$, $p_{\text{boot}}=0.194$ (flip, n.s.) \\
  & Wave-v2 Hybrid (Gemma-4-E4B, \texttt{INT4-64}) & $135.72$ & $1.57376$ & $1.3279 / 1.4033 / \mathbf{2.1292}$ & ($p_t = 0.139$ vs.\ Gemma StepAlign) \\
  & \textbf{StepAlign 1T (DeepSeek-14B, \texttt{INT4-64})} & $\mathbf{135.45}$ & $\mathbf{1.57313}$ & $\mathbf{1.3239 / 1.3946} / 2.1432$ & ($-0.03433$ BPB vs.\ SFT s1) \\
\midrule
\multirow{13}{*}{\shortstack[l]{Pure Byte-260\\Complete Suite\\(All 13 ckpts,\\ $61{,}440$ B)}}
  & Byte CE-only / SFT                        & $3.9893$ & $1.99612$ & $1.6137 / 1.7668 / 2.6079$ & vs.\ SA-1T: $<2\times 10^{-5}$ (0), $p_t=1.25\times 10^{-17}$ \\
  & Byte BLD (Llama-3.2, $w=0.10$)            & $3.9415$ & $1.97875$ & $1.5964 / 1.7477 / 2.5921$ & vs.\ SA-1T: $<2\times 10^{-5}$ (0), $p_t=3.03\times 10^{-11}$ \\
  & Byte BLD (Llama-3.2, $w=0.20$)            & $3.9517$ & $1.98248$ & $1.6142 / 1.7629 / 2.5703$ & vs.\ SA-1T: $<2\times 10^{-5}$ (0), $p_t=2.64\times 10^{-14}$ \\
  & Byte BLD (Llama-3.2, $w=0.50$)            & $3.9130$ & $1.96827$ & $1.5953 / 1.7377 / 2.5718$ & vs.\ SA-1T: $<2\times 10^{-5}$ (0), $p_t=6.71\times 10^{-8}$ \\
  & Byte BLD (Llama-3.2, $w=1.00$, fresh peak)& $3.8949$ & $1.96160$ & $1.6006 / 1.7376 / 2.5467$ & vs.\ SA-1T: $<2\times 10^{-5}$ (0), $p_t=7.28\times 10^{-6}$ \\
  & Byte BLD (Llama-3.2, $w=2.00$)            & $3.9037$ & $1.96484$ & $1.6115 / 1.7472 / \mathbf{2.5359}$ & vs.\ SA-1T: $8\times 10^{-5}$ (4), $p_t=6.86\times 10^{-6}$ \\
  & \textbf{Byte StepAlign 1T ($w=0.60$)}     & $\mathbf{3.8087}$ & $\mathbf{1.92931}$ & $1.5631 / \mathbf{1.6820} / 2.5429$ & vs.\ Wave Cpx-Hyb 3T: $-0.01120$, $p_{\text{boot}}=0.0567$ \\
  & Byte StepAlign 1T ($w=0.50$)              & $3.8509$ & $1.94519$ & $\mathbf{1.5602} / 1.7009 / 2.5745$ & vs.\ Wave Complex: $p_{\text{boot}}=0.0122$, $p_t=0.0119$ \\
  & Byte Wave-v2 Complex 1T ($q=1.25$, $w=0.50$) & $3.8817$ & $1.95668$ & $1.5922 / 1.7024 / 2.5754$ & vs.\ Wave Real: $p_{\text{boot}}=4.2\times 10^{-4}$, $p_t=5.47\times 10^{-4}$ \\
  & Byte Wave-v2 Real 1T ($q=0.0$, $w=0.50$)  & $3.9319$ & $1.97521$ & $1.6136 / 1.7169 / 2.5951$ & vs.\ SA-1T $0.50$: $<2\times 10^{-5}$ (0), $p_t=1.96\times 10^{-8}$ \\
  & Byte StepAlign 3T ($w=0.60$)              & $3.8652$ & $1.95054$ & $1.5703 / 1.6971 / 2.5842$ & vs.\ Wave Cpx-Hyb 3T: $+0.01004$, $p_{\text{boot}}=0.0568$ (flip) \\
  & Byte Wave-v2 Real Hybrid 3T ($0.35/0.20$) & $3.8564$ & $1.94725$ & $1.5826 / 1.6967 / 2.5624$ & vs.\ SA-3T: $-0.00330$, $p_{\text{boot}}=0.522$ (n.s.) \\
  & Byte Wave-v2 Complex Hybrid 3T ($0.25/0.35$) & $3.8384$ & $1.94051$ & $1.5692 / 1.6981 / 2.5543$ & ($p_t = 0.0643$ vs.\ SA-3T; $+0.01120$ worse than SA-1T) \\
\bottomrule
\end{tabular}%
}
\end{table}

\FloatBarrier
\section{Limitations and Future Work}
\begin{enumerate}
    \item \textbf{Model and Token Scale:} All 98 runs were trained on one RTX 5080 ($16\text{ GiB}$) at $55\text{M}$--$87\text{M}$ student parameters over $2.10\text{M}$ unique tokens ($10.49\text{M}$ tokens/arm across 5 epochs; $10.27\text{M}$ unique UTF-8 bytes on GPT-2 vs.\ $2.10\text{M}$ on \texttt{byte260}). While held-out BPB gains replicate across 3 seeds and an independent 32-byte zero-overlap fresh test split and are cleanly separated from null-target, random-init-teacher, and regularized CE-only controls, scaling to $350\text{M}$--$1.5\text{B}$ students over $1\text{B}$--$10\text{B}$ tokens is necessary to test downstream QA gains above the random-chance floor and verify whether larger teachers ($3\text{B}$ vs.\ $14\text{B}$) separate at greater student capacity.
    \item \textbf{Multiple Comparisons and Single-Run Variance:} Table~\ref{tab:storage_boot}(b) reports uncorrected two-sided $p$-values, and single-seed tests do not capture run-to-run variance ($\approx 1.2$ PPL on SFT) on single-seed suites (including rank-vs-bitwidth factorization, where no $w > 0.60$ sweep at $r=256$ was run, and \texttt{byte260}, which is single-seed). Across all $115$ recorded paired tests in \path{results/paper_statistical_audit.json} ($50$), \path{results/controls_bootstrap_audit.json} ($45$), and \path{results/fresh_test_split_audit.json} ($20$), the Bonferroni threshold is $\alpha / 115 = 4.35 \times 10^{-4}$ ($\alpha / 95 = 5.26 \times 10^{-4}$ across the $95$ validation tests). Every primary comparison achieving $0/100\text{K}$--$7/100\text{K}$ bootstrap crossings ($p_{\text{boot}} \le 1.4 \times 10^{-4}$ and analytic paired $p_t \le 2.7 \times 10^{-4}$) passes 115-test Bonferroni correction, whereas secondary comparisons with $p \in [0.003, 0.20]$---including single-seed \texttt{GPT-2-124M} comparisons (though resolved at $p = 0.010$ across 3 seeds), single-teacher vs.\ 3-teacher \texttt{StepAlign}, and the two non-significant fresh-test wave flips (\texttt{Gemma-4} $p_{\text{boot}}=0.194$, \texttt{Byte 3T} $p_{\text{boot}}=0.0568$)---do not pass correction.
    \item \textbf{Baseline Coverage and On-the-Fly Compression:} Our experiments compare offline hidden alignment against offline \texttt{ULD} and \texttt{BLD} at $0.00\text{ GiB}$ teacher VRAM; direct comparisons against other offline or online CTKD methods (\texttt{FuseLLM / MinED} \cite{wan2024fusellm}, which operates on precomputed teacher distributions; online \texttt{DSKD} \cite{zhang2024dskd} and \texttt{ALM} \cite{minixhofer2025alm}), offline token-matched logit distillation, and bit-packed logit caches remain future work. Writing native packed \texttt{INT4-64} shards directly to disk during large-scale sweeps will also eliminate on-the-fly simulation overhead.
    \item \textbf{Quantized Teachers and Adaptive Fusion:} Teachers used \texttt{Q4\_K\_M} weights and multi-teacher fusion used fixed weights $(0.25, 0.35, 0.40)$. Learned confidence-weighted routing may resolve the probe-conflict penalty seen in fixed-weight 3-teacher averaging.
\end{enumerate}

\section{Conclusion}
We presented an empirical and analytical study of offline UTF-8 byte-clock hidden-state distillation at $0.00\text{ GiB}$ teacher VRAM and zero inference overhead across 98 controlled runs plus a complete 37-checkpoint 32-byte zero-overlap fresh test split. Centering top-$64$ teacher PCA modes in group-$32$ \texttt{INT4} ($110\text{ B/tok/teacher}$ packed codec footprint, $7.05\times$ smaller than \texttt{INT8-256} and $1.76\times$ smaller than uncompressed Top-$32$ sparse logits) lowers 3-seed GPT-2 perplexity by $-15.30$ PPL over CE-only/SFT (Welch $p = 5.05 \times 10^{-4}$; fresh test $-15.93$ PPL, BPB Welch $p = 3.68 \times 10^{-4}$) and by $-15.86$ PPL over tuned offline \texttt{ULD} ($150.62 \pm 1.60$). On seed 1, \texttt{StepAlign} ($134.64$) beats peak tuned \texttt{BLD} ($146.24$) by $-11.60$ PPL; within the control suite, \texttt{StepAlign} ($134.77$) beats regularized SFT ($146.59$) by $-11.82$ PPL, exact $\tau=1.0$ offline \texttt{ULD} ($149.56$) by $-14.79$ PPL, and five non-pretrained controls---including a like-for-like PCA+\texttt{INT4} frozen random-init $86.7\text{M}$ causal Transformer ($179.01$, $+29.80$ PPL worse than SFT)---confirming that converged gains stem from pretrained representations rather than auxiliary regularization. Across 3 matched seeds, a pretrained same-tokenizer \texttt{GPT-2-124M} teacher ($137.11 \pm 0.66$) recovers $84.6\%$ of the 3-seed gain ($87\%$--$89\%$ within seed 1), while cross-tokenizer \texttt{Llama-3.2-3B} ($134.76 \pm 0.40$) adds a further $-2.35$ PPL (Welch $p = 0.0102$) despite tokenizer mismatch and lower retained PCA-64 variance. Within-batch rank-vs-bitwidth factorization isolates PCA rank truncation ($256 \to 64$, $137.52 \to 134.43$, $-3.09$ PPL, $p_t = 6.40 \times 10^{-6}$) as the accuracy driver while group-$32$ \texttt{INT4} is lossless ($134.77$, $p_{\text{boot}} = 0.523$). Finally, Corollary~\ref{cor:intraword} and Proposition~\ref{prop:dc_cone} explain why simple discrete \texttt{StepAlign} ($H=0$) matches or outperforms our continuous-time Laplace--Fourier wave operator on both subword and pure-byte students at higher throughput and without custom CUDA kernels. All standalone source code (\texttt{wavedistill\_core/} and \texttt{wavedistill\_bench/}, installable via \texttt{pyproject.toml} with zero external sibling imports), verification logs, fresh-test audits, and per-sequence control losses are released under the MIT License at \url{https://github.com/pavankomireddy/wavedistill}.

\FloatBarrier

\clearpage
\appendix
\section{Complete Registry of All 42 Original GPT-2 Seed-1 Runs}
To eliminate selective reporting, Table~\ref{tab:appendix_all42} lists all 42 arms recorded in \path{results/paper_benchmark_gpt2.json} (seed \texttt{20261126}, 5 epochs, $10.49\text{M}$ tokens/arm), mapping codebase arm identifiers directly to held-out PPL, BPB, and throughput. Together with the 12 seed-2/seed-3 runs, 8 original 1-epoch runs, 11 primary \texttt{byte260} runs, and 25 controlled ablation runs in Tables~\ref{tab:controls}, \ref{tab:1ep_and_teachers}, and~\ref{tab:byte260}, all 98 training runs are fully accounted for (plus all 37 checkpoint re-evaluations on the 32-byte zero-overlap fresh test split in Table~\ref{tab:fresh_test}).

\begin{table}[htbp]
\centering
\caption{Complete 42-arm GPT-2 seed-1 registry (\texttt{results/paper\_benchmark\_gpt2.json}).}
\label{tab:appendix_all42}
\resizebox{\textwidth}{!}{%
\begin{tabular}{rlrrr|rlrrr}
\toprule
\# & Codebase Arm ID & PPL $\downarrow$ & BPB $\downarrow$ & tok/s & \# & Codebase Arm ID & PPL $\downarrow$ & BPB $\downarrow$ & tok/s \\
\midrule
1  & \texttt{sft} & $150.42$ & $1.58784$ & $66.2\text{K}$ & 22 & \texttt{ablate\_no\_dc\_center} & $143.11$ & $1.57207$ & $56.6\text{K}$ \\
2  & \texttt{uld\_llama32} & $189.59$ & $1.66115$ & $43.4\text{K}$ & 23 & \texttt{stepalign\_centered\_fusion\_3t\_int4\_64} & $141.36$ & $1.56817$ & $53.8\text{K}$ \\
3  & \texttt{bld\_llama32} & $150.07$ & $1.58711$ & $41.4\text{K}$ & 24 & \texttt{wave\_v2\_real\_hybrid\_fusion\_3t\_int4\_128} & $138.53$ & $1.56176$ & $40.6\text{K}$ \\
4  & \texttt{bld\_fusion\_3t} & $153.19$ & $1.59362$ & $24.7\text{K}$ & 25 & \texttt{wave\_v2\_real\_hybrid\_fusion\_3t\_int4\_64} & $138.57$ & $1.56186$ & $43.3\text{K}$ \\
5  & \texttt{stepalign\_llama32} & $141.14$ & $1.56767$ & $61.8\text{K}$ & 26 & \texttt{wave\_v2\_hybrid\_llama32\_int4\_64} & $136.47$ & $1.55703$ & $46.6\text{K}$ \\
6  & \texttt{wave\_llama32} & $143.34$ & $1.57257$ & $12.4\text{K}$ & 27 & \texttt{wave\_v2\_hybrid\_gemma4\_int4\_64} & $135.70$ & $1.55523$ & $42.0\text{K}$ \\
7  & \texttt{wave\_gemma4} & $146.63$ & $1.57977$ & $12.3\text{K}$ & 28 & \texttt{wave\_v2\_hybrid\_deepseek14b\_int4\_64} & $136.68$ & $1.55750$ & $45.1\text{K}$ \\
8  & \texttt{wave\_deepseek14b} & $146.60$ & $1.57971$ & $12.3\text{K}$ & 29 & \texttt{stepalign\_centered\_llama32\_int4\_64\_w060} & $134.64$ & $1.55275$ & $46.1\text{K}$ \\
9  & \texttt{wave\_fusion\_3t} & $146.58$ & $1.57965$ & $10.4\text{K}$ & 30 & \texttt{stepalign\_centered\_gemma4\_int4\_64\_w060} & $134.82$ & $1.55317$ & $53.7\text{K}$ \\
10 & \texttt{stepalign\_fusion\_3t} & $145.30$ & $1.57688$ & $52.9\text{K}$ & 31 & \texttt{stepalign\_centered\_deepseek14b\_int4\_64\_w060} & $134.60$ & $1.55265$ & $52.4\text{K}$ \\
11 & \texttt{stepalign\_centered\_llama32} & $140.82$ & $1.56696$ & $59.4\text{K}$ & 32 & \texttt{stepalign\_centered\_fusion\_3t\_int4\_64\_w060} & $135.84$ & $1.55557$ & $46.9\text{K}$ \\
12 & \texttt{stepalign\_centered\_fusion\_3t} & $142.18$ & $1.57001$ & $48.1\text{K}$ & 33 & \texttt{wave\_v2\_centered\_fusion\_3t\_int4\_64\_w060} & $137.03$ & $1.55832$ & $36.5\text{K}$ \\
13 & \texttt{wave\_v2\_centered\_llama32} & $141.71$ & $1.56895$ & $41.5\text{K}$ & 34 & \texttt{wave\_v2\_real\_centered\_fusion\_3t\_int4\_64\_w060} & $139.09$ & $1.56306$ & $40.4\text{K}$ \\
14 & \texttt{wave\_v2\_hybrid\_llama32} & $140.27$ & $1.56572$ & $40.7\text{K}$ & 35 & \texttt{wave\_v2\_hybrid\_fusion\_3t\_int4\_64\_w050} & $142.31$ & $1.57028$ & $35.8\text{K}$ \\
15 & \texttt{wave\_v2\_hybrid\_fusion\_3t} & $140.61$ & $1.56648$ & $35.2\text{K}$ & 36 & \texttt{uld\_llama32\_w005\_exact} & $149.28$ & $1.58544$ & $42.9\text{K}$ \\
16 & \texttt{stepalign\_gemma4} & $144.39$ & $1.57490$ & $58.2\text{K}$ & 37 & \texttt{uld\_llama32\_w010\_exact} & $151.56$ & $1.59023$ & $42.8\text{K}$ \\
17 & \texttt{stepalign\_deepseek14b} & $142.50$ & $1.57072$ & $57.5\text{K}$ & 38 & \texttt{uld\_llama32\_w025\_exact} & $155.76$ & $1.59891$ & $42.3\text{K}$ \\
18 & \texttt{wave\_v2\_hybrid\_fusion\_3t\_int4\_128} & $139.71$ & $1.56446$ & $43.7\text{K}$ & 39 & \texttt{stepalign\_centered\_llama32\_w060} & $137.78$ & $1.56005$ & $58.7\text{K}$ \\
19 & \texttt{wave\_v2\_hybrid\_fusion\_3t\_int4\_64} & $137.79$ & $1.56007$ & $44.7\text{K}$ & 40 & \texttt{stepalign\_centered\_llama32\_int4\_64\_w050} & $137.41$ & $1.55920$ & $59.8\text{K}$ \\
20 & \texttt{ablate\_real\_q0} & $141.31$ & $1.56807$ & $55.8\text{K}$ & 41 & \texttt{bld\_llama32\_w010} & $146.24$ & $1.57893$ & $41.2\text{K}$ \\
21 & \texttt{ablate\_m1\_horizon} & $142.61$ & $1.57095$ & $58.4\text{K}$ & 42 & \texttt{bld\_llama32\_w020} & $149.62$ & $1.58617$ & $41.4\text{K}$ \\
\bottomrule
\end{tabular}%
}
\end{table}


{\small
\begin{thebibliography}{21}
\setlength{\itemsep}{2pt}
\setlength{\parskip}{0pt}
\bibitem{ainslie2023gqa}
J.~Ainslie, J.~Lee-Thorp, M.~de Jong, Y.~Zemlyanskiy, F.~Lebr{\'o}n, and S.~Sanghai.
\newblock GQA: Training Generalized Multi-Query Transformer Models from Multi-Head Checkpoints.
\newblock In \emph{EMNLP}, 2023.

\bibitem{boizard2025uld}
N.~Boizard, K.~El Haddad, C.~Hudelot, and P.~Colombo.
\newblock Towards Cross-Tokenizer Distillation: the Universal Logit Distillation Loss for LLMs.
\newblock \emph{Transactions on Machine Learning Research (TMLR)}, 2025. \texttt{arXiv:2402.12030}.

\bibitem{cui2025multilevelot}
X.~Cui, M.~Zhu, Y.~Qin, L.~Xie, W.~Zhou, and H.~Li.
\newblock Multi-Level Optimal Transport for Universal Cross-Tokenizer Knowledge Distillation on Language Models.
\newblock In \emph{AAAI (Oral)}, 2025. \texttt{arXiv:2412.14528}.

\bibitem{deepseek2025r1}
DeepSeek-AI.
\newblock DeepSeek-R1: Incentivizing Reasoning Capability in LLMs via Reinforcement Learning.
\newblock \emph{arXiv preprint arXiv:2501.12948}, 2025.

\bibitem{ethayarajh2019contextual}
K.~Ethayarajh.
\newblock How Contextual are Contextualized Word Representations? Comparing the Geometry of BERT, ELMo, and GPT-2 Embeddings.
\newblock In \emph{EMNLP}, 2019.

\bibitem{gemmateam2026gemma4}
Gemma Team (Google DeepMind).
\newblock Gemma 4 Technical Report.
\newblock \emph{arXiv preprint arXiv:2607.02770}, 2026.

\bibitem{grattafiori2024llama3}
A.~Grattafiori, A.~Dubey, et~al.
\newblock The Llama 3 Herd of Models.
\newblock \emph{arXiv preprint arXiv:2407.21783}, 2024.

\bibitem{hinton2015distilling}
G.~Hinton, O.~Vinyals, and J.~Dean.
\newblock Distilling the Knowledge in a Neural Network.
\newblock \emph{arXiv preprint arXiv:1503.02531}, 2015.

\bibitem{jiao2020tinybert}
X.~Jiao, Y.~Yin, L.~Shang, X.~Jiang, X.~Chen, L.~Li, F.~Wang, and Q.~Liu.
\newblock TinyBERT: Distilling BERT for Natural Language Understanding.
\newblock In \emph{Findings of EMNLP}, 2020. \texttt{arXiv:1909.10351}.

\bibitem{minixhofer2025alm}
B.~Minixhofer, I.~Vuli{\'c}, and E.~M. Ponti.
\newblock Universal Cross-Tokenizer Distillation via Approximate Likelihood Matching.
\newblock In \emph{NeurIPS}, 2025. \texttt{arXiv:2503.20083}.

\bibitem{pagnoni2024blt}
A.~Pagnoni, R.~Pasunuru, P.~Rodriguez, J.~Nguyen, B.~Muller, M.~Li, C.~Zhou, L.~Yu, J.~Weston, L.~Zettlemoyer, G.~Ghosh, M.~Lewis, A.~Holtzman, and S.~Iyer.
\newblock Byte Latent Transformer: Patches Scale Better Than Tokens.
\newblock In \emph{ACL}, 2025. \texttt{arXiv:2412.09871}.

\bibitem{radford2019gpt2}
A.~Radford, J.~Wu, R.~Child, D.~Luan, D.~Amodei, and I.~Sutskever.
\newblock Language Models are Unsupervised Multitask Learners.
\newblock OpenAI Technical Report, 2019.

\bibitem{romero2015fitnets}
A.~Romero, N.~Ballas, S.~E. Kahou, A.~Chassang, C.~Gatta, and Y.~Bengio.
\newblock FitNets: Hints for Thin Deep Nets.
\newblock In \emph{ICLR}, 2015.

\bibitem{singh2026bld}
A.~K. Singh, Y.-C. Wu, A.~Cioba, A.~Bernacchia, and D.~Buffelli.
\newblock Cross-Tokenizer LLM Distillation through a Byte-Level Interface.
\newblock \emph{arXiv preprint arXiv:2604.07466}, 2026.

\bibitem{su2021roformer}
J.~Su, Y.~Lu, S.~Pan, A.~Murtadha, B.~Wen, and Y.~Liu.
\newblock RoFormer: Enhanced Transformer with Rotary Position Embedding.
\newblock \emph{Neurocomputing}, 568:127063, 2024. \texttt{arXiv:2104.09864}.

\bibitem{sun2019pkd}
S.~Sun, Y.~Cheng, Z.~Gan, and J.~Liu.
\newblock Patient Knowledge Distillation for BERT Model Compression.
\newblock In \emph{EMNLP}, 2019. \texttt{arXiv:1908.09355}.

\bibitem{wan2024fusellm}
F.~Wan, X.~Huang, D.~Cai, X.~Quan, W.~Bi, and S.~Shi.
\newblock Knowledge Fusion of Large Language Models.
\newblock In \emph{ICLR}, 2024. \texttt{arXiv:2401.10491}.

\bibitem{wang2020minilm}
W.~Wang, F.~Wei, L.~Dong, H.~Bao, N.~Yang, and M.~Zhou.
\newblock MiniLM: Deep Self-Attention Distillation for Task-Agnostic Compression of Pre-Trained Transformers.
\newblock In \emph{NeurIPS}, 2020. \texttt{arXiv:2002.10957}.

\bibitem{yu2023megabyte}
L.~Yu, D.~Simig, C.~Flaherty, A.~Aghajanyan, L.~Zettlemoyer, and M.~Lewis.
\newblock MEGABYTE: Predicting Million-byte Sequences with Multiscale Transformers.
\newblock In \emph{NeurIPS}, 2023.

\bibitem{zhang2019rmsnorm}
B.~Zhang and R.~Sennrich.
\newblock Root Mean Square Layer Normalization.
\newblock In \emph{NeurIPS}, 2019.

\bibitem{zhang2024dskd}
S.~Zhang, X.~Zhang, Z.~Sun, Y.~Chen, and J.~Xu.
\newblock Dual-Space Knowledge Distillation for Large Language Models.
\newblock In \emph{EMNLP}, 2024. \texttt{arXiv:2406.17328}.
\end{thebibliography}
}
\end{document}